\documentclass{article} %
\usepackage{collas2026_conference,times}
\usepackage{amsmath,amsfonts,bm}

\def\eqref#1{equation~\ref{#1}}

\def\1{\bm{1}}

\DeclareMathAlphabet{\mathsfit}{\encodingdefault}{\sfdefault}{m}{sl}
\SetMathAlphabet{\mathsfit}{bold}{\encodingdefault}{\sfdefault}{bx}{n}

\usepackage{hyperref}
\hypersetup{
    colorlinks=true,
    linkcolor=red,
    filecolor=magenta,
    urlcolor=blue,
    citecolor=purple,
    pdftitle={TOOD Task-Aware Out-of-Distribution Score Calibration for Continual Learners},
    pdfpagemode=FullScreen,
    }

\usepackage{booktabs}
\usepackage{algorithm}
\usepackage[noend]{algpseudocode}
\usepackage{subcaption}
\usepackage{graphicx}   
\usepackage{placeins}
\usepackage{enumitem}
\usepackage{xcolor}
\definecolor{teal}{rgb}{0.0, 0.5, 0.5}
\usepackage{acronym}

\acrodef{OOD}{out-of-distribution}
\acrodef{TOOD}{Task-Aware OOD Score Calibration}
\acrodef{ID}{in-distribution}
\acrodef{AUROC}{area under the receiver operating characteristic curve}
\acrodef{DNN}{deep neural network}
\acrodef{CL}{Continual Learning}
\acrodef{CIL}{Class-Incremental Learning} 
\acrodef{BER}{Bi-directional Energy Regularization}
\acrodef{MAD}{Median Absolute Deviation}
\acrodef{CKA}{Centered Kernel Alignment}
\acrodef{TIL}{Task-Incremental Learning} 
\acrodef{MLP}{Multi-Layer Perceptron}

\usepackage{cleveref}  
\crefname{app}{Appendix}{Appendices}
\Crefname{app}{Appendix}{Appendices}

\title{TOOD: Task-Aware Out-of-Distribution Score Calibration for Continual Learners}

\author{Mostafa ElAraby \\
DIRO, Mila - Quebec AI Institute \\
Universit\'e de Montr\'eal\\
Montreal, Canada \\
\texttt{\{elarabim\}@mila.quebec} \\
\And %
Samer B. Nashed  \\
DIRO, Mila - Quebec AI Institute \\
Universit\'e de Montr\'eal\\
Montreal, Canada \\
\AND
Liam Paull  \\
CIFAR AI Chair\\
DIRO, Mila - Quebec AI Institute \\
Universit\'e de Montr\'eal\\
Montreal, Canada \\
}

\preprintcopy %

\begin{document}

\maketitle

\begin{abstract}
    The primary challenge of continual learning (CL) systems is to learn new tasks while remaining performant on previously learned tasks. A similarly important though less well-studied aspect of CL systems is their ability to identify inputs that are unlikely to come from within the set of tasks the system has already encountered, often called out-of-distribution (OOD) detection. This paper presents several findings related to the dynamics of OOD detection in CL systems, causes of performance degradation over time which we call \emph{OOD forgetting} (OODF), and proposed mitigation strategies for OODF. Chiefly, we find the unintuitive result that OODF is only weakly anti-correlated with classification performance on previous tasks, suggesting that the underlying mechanisms producing OODF are distinct. Moreover, this effect is observed for both energy-based and feature-based OOD detection methods. Energy-based detectors suffer a drop in logit scale as additional tasks are learned, which we term the \emph{Confidence Gap}, while feature-based detectors also degrade under a complementary effect we call \emph{Manifold Crowding}. Motivated by these observations, we propose \ac{TOOD}, a training-free post-hoc method that decomposes logits into per-task energy scores and re-calibrates them using replay-buffer statistics. Experiments on CIFAR-10, CIFAR-100, and a 100-task ImageNet-1K stream show that TOOD improves OOD detection performance over uncalibrated energy in most settings and ranks first or second in eight of ten CIFAR configurations under 3-seed mean$\pm$std evaluation, with the largest gains when the confidence gap is most severe. These results suggest that a portion of OOD deterioration in continual learning arises from score miscalibration rather than from a complete loss of discriminative structure.

\end{abstract}

\section{Introduction}
Reliable deployment of a \ac{CL} system requires two abilities: 1) retaining performance on previously learned tasks and 2) recognizing when a test input falls outside the set of learned classes. The \ac{CL} literature has made substantial progress on the first objective through regularization~\citep{kirkpatrick2017overcoming}, replay~\citep{rebuffi2017icarl,chaudhry2019tiny}, custom architectures~\citep{rusu2016progressive,serra2018overcoming} and distillation-based methods~\citep{buzzega2020dark,li2017learning}. The second objective, robust \ac{OOD} detection across the learning stream, has received much less direct attention.

Robust \ac{OOD} detection requires stable data distributions and calibrated output scores~\citep{hendrycks2017baseline,yang2021generalized}, assumptions that \ac{CL} violates through continuous parameter updates. \ac{CL} systems further exacerbate these challenges via two phenomena. First, the semantic boundary of what constitutes \ac{OOD} data is a moving target: as new tasks are introduced, regions of the input space previously considered unknown are absorbed into the \ac{ID} manifold. Second, as the model accommodates these new classes, the classifier head continuously expands, permanently altering the absolute scale and geometry of the output space. Such OOD separability degradation represents a silent, critical failure mode in continual learning (CL) systems. While standard metrics evaluate how well a system remembers past classes, they fail to capture its ability to identify novel, unseen data—a fundamental safety requirement for the responsible real-world deployment of lifelong learning systems.

This leads to the following research questions: \textbf{RQ1:} Are declines in OOD detection and classifier accuracy driven by different underlying mechanisms? \textbf{RQ2:} What underlying mechanisms drive the progressive deterioration of \ac{OOD} detection during \ac{CL}? \textbf{RQ3:} Can this deterioration be mitigated via a training-free, post-hoc intervention?

To address these questions in light of the above challenges, we first measure out-of-distribution forgetting (OODF) for various \ac{CL} and \ac{OOD} detection methods in order to uncover the underlying mechanisms driving OODF. Our empirical results demonstrate that OODF is not merely a byproduct of forgetting old tasks, but a distinct failure of the model’s confidence calibration as its output capacity expands. A CL system that appears `well-preserved' in its classification performance can simultaneously become overconfident on novel inputs.
Specifically, our analysis points to two recurring mechanisms behind this deterioration. First, as the \ac{CL} system is presented with new tasks and the classifier head expands, the logits associated with older tasks tend to shrink in absolute scale relative to the current task; we refer to this task-age-dependent score drift as the \emph{confidence gap}. Second, as more classes are introduced, the latent feature space becomes more densely occupied, reducing the margin available to distance-based detectors; we refer to this as \emph{manifold crowding}. Together, these observations suggest that \ac{OOD} forgetting is not fully characterized by standard accuracy-based forgetting metrics (see~\Cref{sec:problem}). 
Existing evaluation protocols measure this degradation only at the final checkpoint, missing how detection erodes over the task stream.

Motivated by this diagnosis, we propose \textbf{Task-Aware OOD Score Calibration (TOOD)}, a training-free, post-hoc method that decomposes logits into per-task energy scores and re-calibrates those scores using task-wise \ac{ID} statistics. 
Requiring only a single forward pass over a small calibration memory and no gradient updates, \ac{TOOD} leaves the underlying \ac{CL} algorithm unmodified and introduces no optimization overhead.

Across several experiments on CIFAR-10 and CIFAR-100, TOOD reduces OODF compared to baselines in the majority of cases and yields its largest gains when score drift is severe and shows less benefit when the baseline already stabilizes output scale, indicating that it is a relatively effective strategy for addressing this phenomenon. Under 3-seed mean$\pm$std evaluation, at least one TOOD variant ranks first or second in eight of the ten CIFAR configurations; the two misses occur where the confidence gap is smallest and the gaps to the next detector are within seed variation.

Overall, this paper presents three contributions. First, we formalize \emph{OOD Forgetting} and introduce metrics that track OOD detection across the full learning trajectory rather than just end-state performance. Second, we identify two mechanisms that produce OODF: the \emph{confidence gap} (task-age logit-scale drift) and \emph{manifold crowding} (progressive latent space saturation), and we demonstrate these recur across a number of state-of-the-art \ac{CL} and \ac{OOD} detection methods. Finally, we propose \textbf{TOOD}, a training-free, post-hoc method that calibrates per-task energy scores using task-wise \ac{ID} calibration statistics, requiring only a single forward pass over that calibration memory and no retraining.

\section{Related Work}
\ac{OOD} detection methods broadly assume a fixed model and a static data distribution~\citep{hendrycks2017baseline, yang2021generalized}. Post-hoc output-based methods assign anomaly scores from model outputs: MSP~\citep{hendrycks2017baseline} uses the maximum softmax probability, ODIN~\citep{liang2017enhancing} applies temperature scaling and input perturbation, and energy scoring~\citep{liu2020energy} aggregates class logits into a scalar. Feature-based methods exploit geometric separation in representation space~\citep{lee2018simple, sun2022out}, while activation-shaping methods such as ReAct and ASH~\citep{sun2021react, djurisic2023extremely} suppress outlier activations before scoring. CL researchers have also explored gradient-based replay constraints~\citep{lopez2017gradient, chaudhry2018efficient}, synaptic consolidation~\citep{zenke2017continual}, and output-layer bias correction or weight alignment~\citep{zhao2019maintaining, wu2019large}. These methods improve old/new class balance for classification, but they are not designed to preserve \ac{OOD} score separation over time or to control the feature-space geometry on which many \ac{OOD} detectors depend. Moreover, all of these approaches assume a frozen backbone and a stable score distribution, which is not the case in continual learning, where parameters update sequentially and the set of known classes grows.

OpenCIL~\citep{miao2024opencil} documents that standard \ac{CL} methods struggle to reject \ac{OOD} samples and benchmarks this problem across incremental steps; CLOM~\citep{kim2022continual} uses \ac{OOD} detection for task inference. A growing line of work studies \ac{OOD} detection and \ac{CL} jointly \cite{gupta2025oodcl}, but most of it adopts a different definition of \ac{OOD} data. \citet{gupta2026bufferfree} use \ac{OOD} scorers for task identity inference and open world recognition so that detection replaces a replay buffer. Separately, \citet{harun2025neuralcollapse} study how neural collapse affects \ac{OOD} detection and generalization. The continual works largely define OOD data as derived from unknown or \emph{future} classes. Our setting is complementary. We define \ac{OOD} as data semantically disjoint from \emph{every} learned task,
and ask how the ability to reject it \emph{degrades along the learning trajectory}, correcting the dominant score level component post hoc. Our work complements OpenCIL by moving from aggregate benchmarking to an explicit analysis of \ac{OOD} forgetting: we formalize task-wise deterioration over the \ac{CL} trajectory and use it to disentangle the mechanisms driving performance loss. In non-\ac{CL} settings, \citet{kumar2022fine} show that full fine-tuning can hurt \ac{OOD} generalization relative to linear probing, while \citet{ming2024does} show that the effect of fine-tuning on \ac{OOD} detection can be strongly score- and method-dependent in vision-language models. We extend that line of inquiry to the continual setting, identifying task-dependent logit-scale drift and feature-space saturation as primary drivers of \ac{OOD} deterioration in our experiments.

Work on score calibration has shown that modern networks are overconfident and temperature scaling improves calibration on i.i.d.\ data~\citep{guo2017calibration}; \citet{minderer2021revisiting} find that calibration degrades under distribution shift. Crucially, any monotonic scalar transformation applied to all test samples preserves score ranking and therefore cannot improve AUROC~\citep{fawcett2006introduction}, motivating TOOD's per-task score decomposition rather than global calibration.

\section{OOD Forgetting}
\label{sec:problem}
Measuring OODF and evaluating potential solutions requires us to formalize the \ac{OOD} detection problem in \ac{CL} systems and then establish several metrics to assess OODF throughout the task stream. First, however, we briefly review some basic \ac{CL} formalism.

Given a model $f_{\theta_t}$ trained on tasks $T_0 \ldots T_t$, we denote the accuracy of  $f_{\theta_t}$ on a test set from task $T_i$ after learning $T_t$ as $A_{t,i}$. Using this measure, task forgetting for task $T_i$ after the learner has trained on $N+1$ tasks is defined as
\begin{equation}
    F_i = \max_{t \in \{i,\ldots,N-1\}} A_{t,i} \;-\; A_{N,i}.
    \label{eq:acc_forgetting}
\end{equation}
Standard \ac{CL} benchmarks such as Split-CIFAR-10~\citep{krizhevsky2009learning}, Split-CIFAR-100~\citep{krizhevsky2009learning}, and Split-ImageNet-1k~\citep{deng2009imagenet} are typically evaluated using $F_i$ or its aggregate $\bar{F} = \frac{1}{N-1}\sum_{i=1}^{N-1} F_i$. This framing implicitly assumes that preserving accuracy also preserves the ability to detect inputs from outside the learned distribution. We show empirically that this assumption does not hold: \Cref{eq:acc_forgetting} captures classification robustness but is insensitive to OOD forgetting, which motivates the metrics we develop next.

Consider a \ac{CL} scenario with disjoint tasks $T_0, \ldots, T_N$. Each task $T_t$ introduces dataset $\mathcal{D}_t = \{(x,y)\}$ from distribution $P_t(x,y)$ over class set $\mathcal{C}_t$, with $\mathcal{C}_i \cap \mathcal{C}_j =  \emptyset$ for $i \neq j$. Let $\theta_t$ denote model parameters after training on $T_t$. After training on $T_N$, the final task, we would like the model to both (1) correctly classify \ac{ID} samples from any seen task, and (2) detect \ac{OOD} samples $\mathcal{D}_{\text{OOD}}$ that possess no semantic overlap with any $\mathcal{C}_t$. Objective~(1) is captured by $F_i$ (\Cref{eq:acc_forgetting}); we now develop a measure for objective~(2) by extending the evaluation framework of OpenCIL~\citep{miao2024opencil} to explicitly quantify task-wise \ac{OOD} forgetting over the full \ac{CL} trajectory.

Let $S: \mathcal{X} \to \mathbb{R}$ be a score function that assigns a real-valued label to data items $x \in \mathcal{X}$. By convention, we interpret scores closer to $-\infty$ to indicate a lower probability of $x$ being an \ac{ID} sample and scores closer to $+\infty$ to indicate a higher probability of $x$ being an \ac{ID} sample. We define the \ac{OOD} detection performance of score function $S(x; \theta_t)$, with respect to task $T_i$ under model $\theta_t$ as $\text{AUROC}(T_i \mid \theta_t) = \Pr[S(x^+;\theta_t) > S(x^-;\theta_t)]$, where $x^+ \sim \mathcal{D}_i$ is a random \ac{ID} sample and $x^- \sim \mathcal{D}_{\text{OOD}}$ is a random \ac{OOD} sample. We define \textbf{\ac{OOD} forgetting} w.r.t.\ task $T_i$ as the drop in \ac{OOD} detection performance between $t=i$ and $t=N$: $D_i = \text{AUROC}(T_i \mid \theta_i) - \text{AUROC}(T_i \mid \theta_N)$.
The average \ac{OOD} forgetting summarizes this over all tasks:
\begin{equation}
    D_{\text{avg}} = \frac{1}{N} \sum_{i=0}^{N-1} D_i.
    \label{eq:avg_det}
\end{equation}
A positive forgetting value, $D_i > 0$, indicates that the model's ability to distinguish between \ac{ID} samples from task $T_i$ and \ac{OOD} samples that are not present in any task has deteriorated since the model was first trained to perform task $T_i$; if $D_i = 0$, performance has remained constant though individual data items may be classified differently; and $D_i < 0$ indicates the model has improved (unlikely). 
$D_i$ is distinct from $F_i$ (\Cref{eq:acc_forgetting}): $F_i$ measures erosion of decision boundaries \emph{between known classes}, while $D_i$ measures erosion of the boundary \emph{between all known classes and the unknown}. Both are driven by sequential gradient updates but respond to different aspects of the model's geometry.

\ac{CL} systems are typically evaluated for \ac{OOD} robustness only at $t=N$, against a fixed \ac{OOD} dataset, which does not allow investigation of the \emph{process} of degradation over sequential learning. Following the step-wise evaluation of \citet{miao2024opencil}, we formalize average incremental AUROC, the OOD analogue of average incremental accuracy~\citep{rebuffi2017icarl}. After learning task $T_t$, the checkpoint $\theta_t$ is evaluated on all tasks seen thus far, $T_0 \ldots T_t$. The per-task AUROC is averaged over the $|\mathcal{O}|$ \ac{OOD} evaluation datasets $\mathcal{O} = \{d_1, \ldots, d_{|\mathcal{O}|}\}$: $\overline{\text{AUROC}}(T_i \mid \theta_t) = \tfrac{1}{|\mathcal{O}|} \sum_{d \in \mathcal{O}} \text{AUROC}_d(T_i \mid \theta_t)$. The aggregate trajectory metric is
\begin{equation}
    \text{Avg\,AUROC} =
        \frac{1}{N+1} \sum_{t=0}^{N}
        \frac{1}{t+1} \sum_{i=0}^{t}
        \overline{\text{AUROC}}(T_i \mid \theta_t).
    \label{eq:avg_auroc}
\end{equation}
We report average incremental AUROC~(\Cref{eq:avg_auroc}), $D_{\text{avg}}$~(\Cref{eq:avg_det}), and false positive rate at 95\% \ac{ID} recall (FPR@95) as complementary metrics.

\subsection{Mechanisms behind OOD Forgetting}

\begin{figure}[t]
    \centering
    \includegraphics[width=\linewidth]%
        {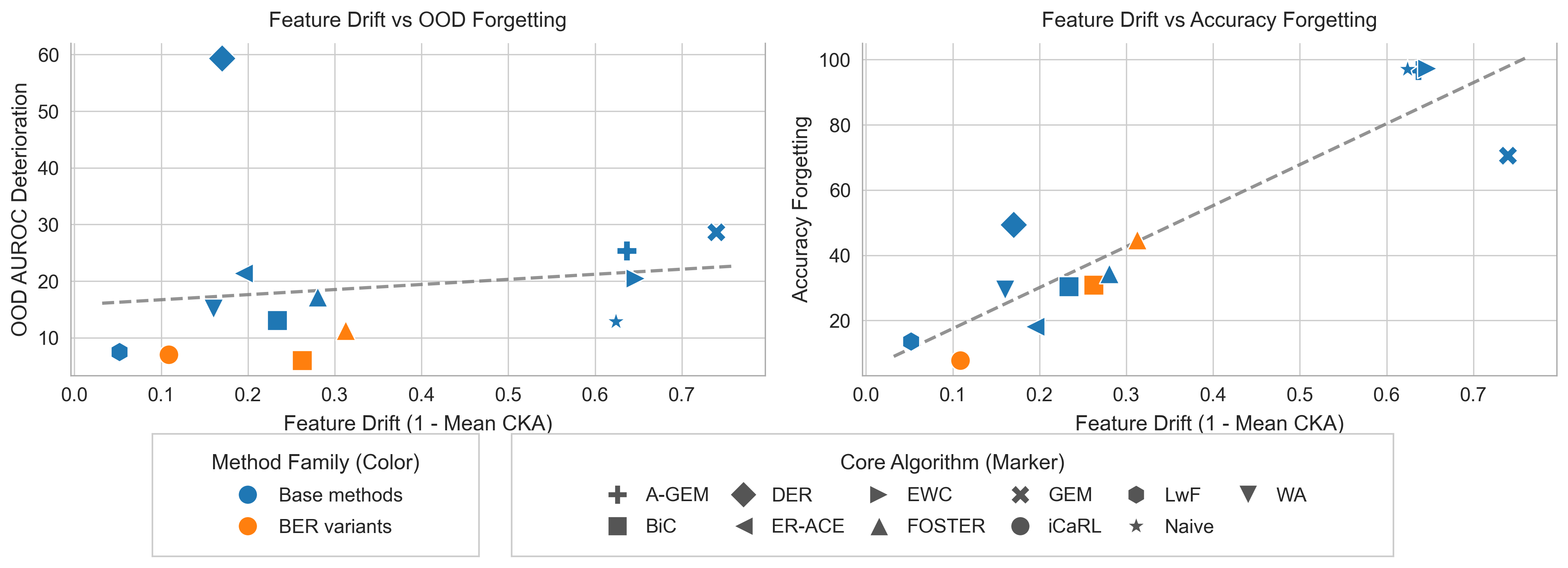}
    \caption{%
        \textbf{Feature drift explains accuracy forgetting but not \ac{OOD} forgetting.} Each point is a \ac{CL} method on CIFAR-10 ($N=5$ tasks). \emph{Left:} Feature drift ($1-\overline{\text{CKA}}$) vs. OOD deterioration ($D_{\text{avg}}$) shows no significant relationship; $r = 0.365$, $p = 0.270$. \emph{Right:} Feature drift vs. classification forgetting ($\bar{F}$); $r = 0.880$, $p < 0.001$. Features drift appears to cause classification forgetting but not OOD detection, suggesting the latter is driven by distinct mechanisms. 
    }
    \label{fig:cka}
\end{figure}

Accuracy forgetting is governed by \emph{decision boundary preservation}, and \ac{CL} methods seek to protect the relative class rankings needed to separate known classes. When they succeed, $A_{N,i} \approx A_{i,i}$ and $F_i \approx 0$. In contrast, all \ac{OOD} detection methods, whether they compute scores from logit energy~\citep{liu2020energy}, feature distances~\citep{lee2018simple, sun2022out}, or activation statistics~\citep{sun2021react, djurisic2023extremely}, rely on \emph{score distribution preservation}. This creates a mismatch between important quantities for \ac{OOD} detection methods and \ac{CL} methods. \Cref{fig:cka} illustrates this mismatch, showing that feature drift, measured using centered kernel alignment (CKA)~\citep{kornblith2019similarity} across \ac{CL} methods on CIFAR-10 ($N=5$), is highly correlated with accuracy forgetting, but much less so with OODF.

This is not surprising considering that under the hood, energy-based \ac{OOD} detection methods operate on the absolute scale of the logit vector, and feature-based detectors rely on unoccupied geometric margins. However, standard \ac{CL} techniques do not constrain these properties. During task acquisition, as the output head expands, old-task logits shrink in absolute scale and the latent space fills with new embeddings. While the relative decision boundary between classes remains intact, preserving classification accuracy, the absolute metrics required to detect \ac{OOD} inputs are altered. Moreover, methods with low feature drift (WA, iCaRL) still exhibit significant OODF, suggesting additional sources of degradation in the output scores, consistent with \citet{ramasesh2021anatomy}, who show that forgetting in deep networks is primarily an output-layer phenomenon.

\paragraph{Manifold Crowding.}
\label{sec:manifold_crowding}

Feature-based detectors such as MDS~\citep{lee2018simple} rely on geometric margin, the distance between an \ac{OOD} sample and the nearest \ac{ID} cluster in representation space. During CL, we observe a degradation mechanism we term \emph{manifold crowding}. As new tasks arrive, their embeddings colonize previously unoccupied latent regions, compressing the margin available for distance-based detection even when task representations remain intact. On a 5-task CIFAR-10 stream, the median nearest-neighbor distance from \ac{OOD} samples to the \ac{ID} manifold decreases by approximately $28\%$ while CKA analysis confirms that feature drift for old tasks remains low (see \Cref{app:manifold_crowding} for figures). This disassociation, shrinking margins without significant representation drift, shows that manifold crowding is driven by the \emph{occupancy} of latent space by new classes rather than by the movement of old ones.
This phenomenon is conceptually related to the superposition hypothesis, where limited representational capacity forces multiple features to share latent directions \citep{elhage2022toy}. In our setting, however, the effect is temporal. Continual class growth progressively occupies regions that previously served as OOD margin, rather than having overlapping features in a jointly trained neural network that are disjoint from OOD representation.

\ac{TOOD} does not directly address manifold crowding. It operates on output logits and cannot restore geometric margin lost in feature space. We confirm this by isolating manifold crowding and the confidence gap in a toy task with fully known geometry (16 Gaussian-blob classes on a sphere, with \ac{OOD} at the interior origin; \Cref{app:abl_toy}). Growing the classifier head while holding the data fixed exposes only the confidence gap, which \ac{TOOD} re-centres for up to $+5.1$ Avg AUROC. However, forcing class overlap to create manifold crowding collapses the geometric margin, which TOOD cannot recover. Moreover, a feature-space analog of \ac{TOOD} that applies the same task-wise normalization in distance space rather than logit space fails to recover separability ($-0.76$ Avg AUROC), since re-centring scores cannot restore margin that newly added classes have already consumed (\Cref{app:abl_crowding}). Manifold crowding therefore represents a distinct and open challenge: improving \ac{OOD} detection for feature-based methods in \ac{CL} systems requires complementary techniques such as representation regularization or margin-preserving replay objectives. Additional results on the 100-task ImageNet-1K stream are reported in \Cref{app:imagenet_scalability}.

\paragraph{The Confidence Gap.}
\label{sec:gap}

The \emph{confidence gap} is the dominant mechanism behind \ac{OOD} forgetting for energy-based detectors. As new tasks arrive, gradient updates prioritize the current task's classes, causing old-task logits to shrink in absolute scale relative to newer tasks. Replay methods illustrate this clearly: rehearsing $k$ exemplars per class preserves the relative ordering of old-task logits but imposes no constraint on the absolute magnitude of the expanding logit vector. Gradient updates for new classes push current-task logits to higher scale, creating a confidence gap even when old-task accuracy is preserved. The confidence gap is summarized by the expression $\mathbb{E}_{x \sim \mathcal{D}_i}[S(x;\theta_N)] \ll \mathbb{E}_{x \sim \mathcal{D}_j}[S(x;\theta_N)]$ for $i < j$, even when $\mathbb{E}_{x \sim \mathcal{D}_i}[A(x;\theta_N)] \approx \mathbb{E}_{x \sim \mathcal{D}_i}[A(x;\theta_i)]$.
\begin{figure*}[t]
    \centering
    \begin{subfigure}[t]{0.49\linewidth}
        \centering
        \includegraphics[width=\linewidth]%
            {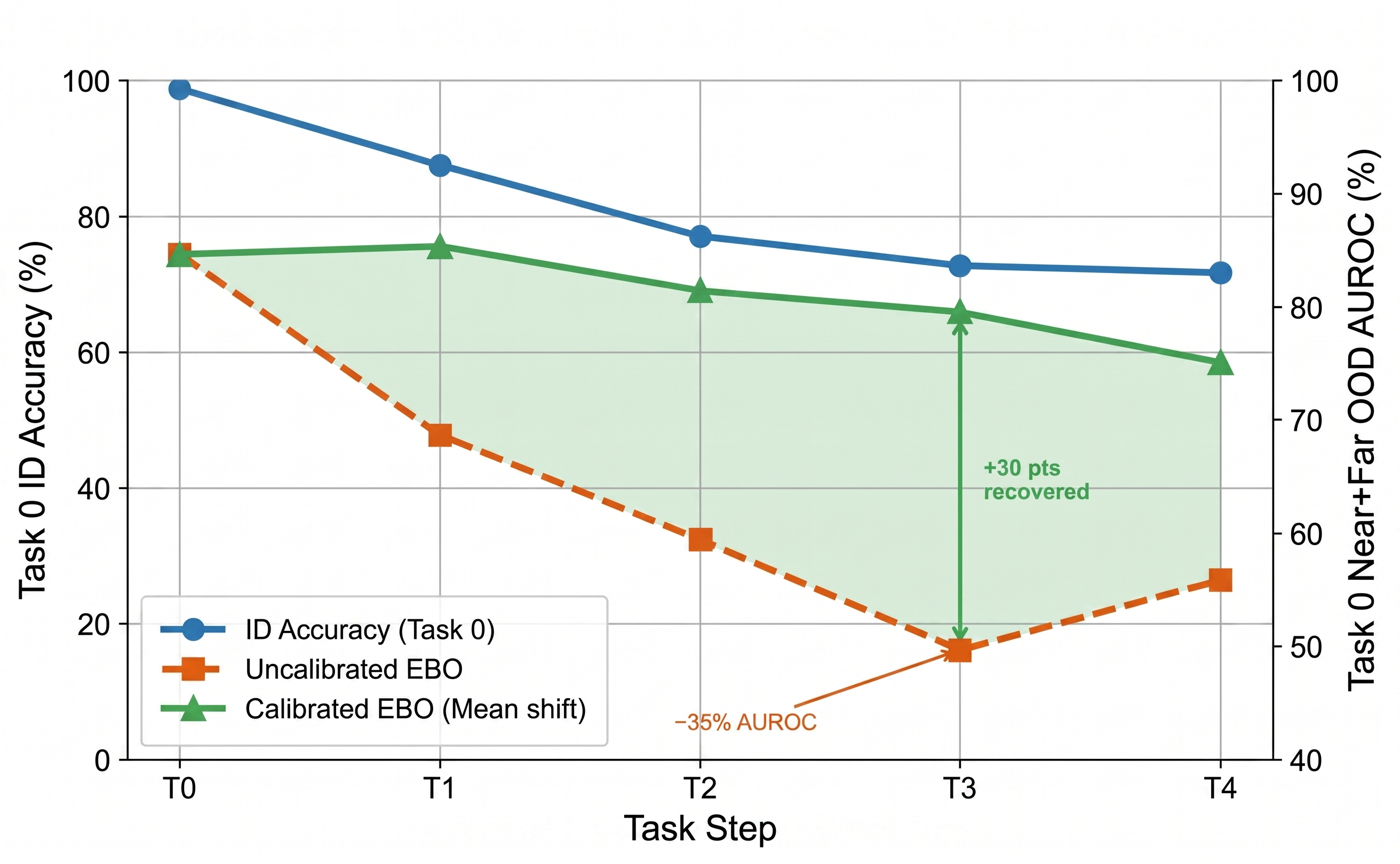}
    \end{subfigure}
    \hfill
    \begin{subfigure}[t]{0.49\linewidth}
        \centering
        \includegraphics[width=\linewidth]%
            {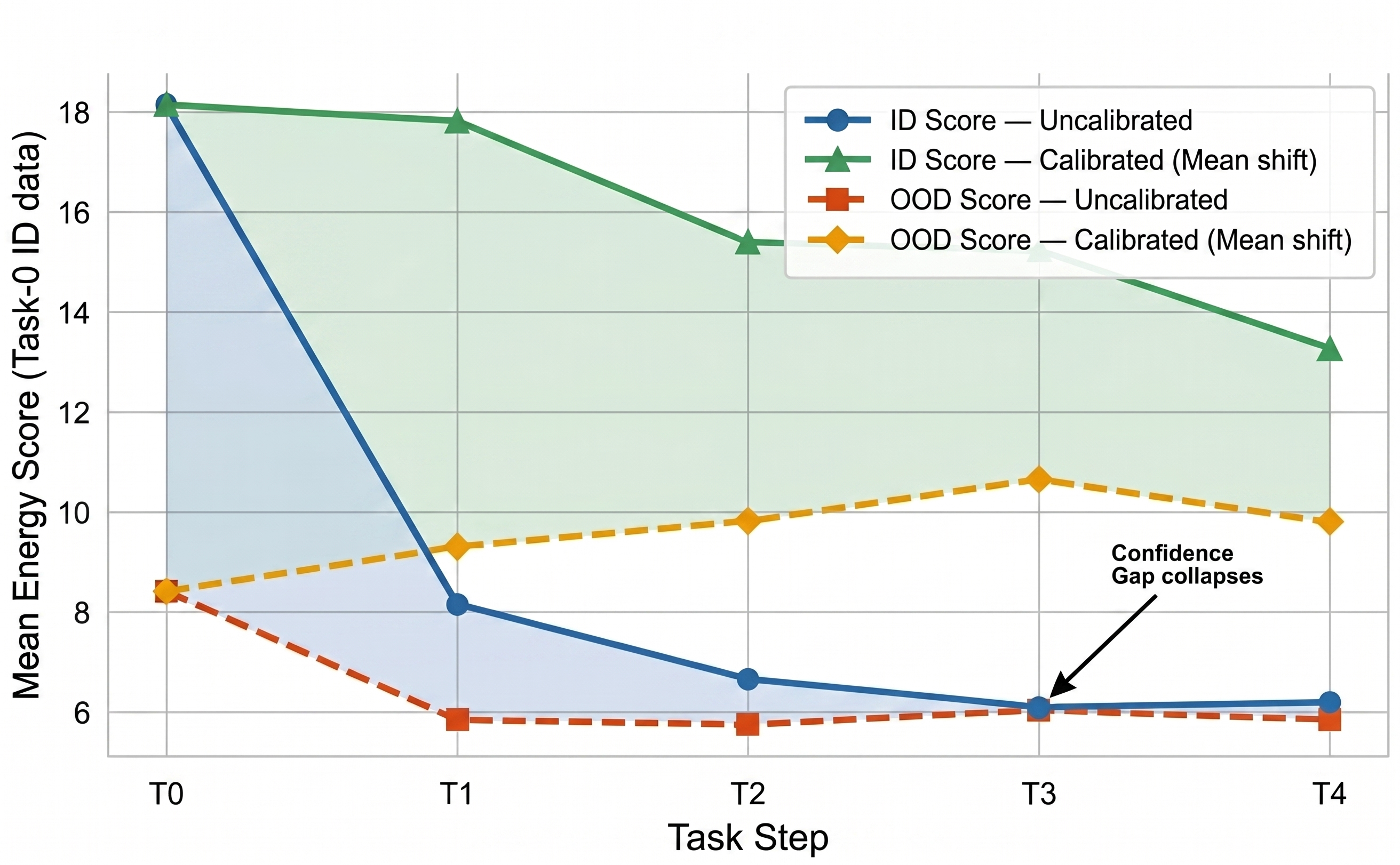}
    \end{subfigure}
    \caption{%
        \textbf{Continual learners retain classification accuracy on old tasks while losing \ac{OOD} separability.}  \emph{Left:} Task~0 classifier acc. (blue) remains high, while uncalibrated energy-based \ac{OOD} AUROC (red dash) drops by $35$ points. \ac{TOOD} calibration (green) recovers most of that gap. \emph{Right:} The mean \ac{ID} energy score (blue) drifts from its $T_0$ value toward the \ac{OOD} score (red dash), collapsing the \textit{confidence gap} at $T_3$. \ac{TOOD} (green) re-centers the task-specific score distribution throughout the stream. Both figures show iCaRL on a 5-task CIFAR-10 stream, tested on Task~0.\looseness-1
    }
    \label{fig:main_panel}
\end{figure*}
\begin{figure}[!ht]
    \centering
    \includegraphics[width=0.9\linewidth]{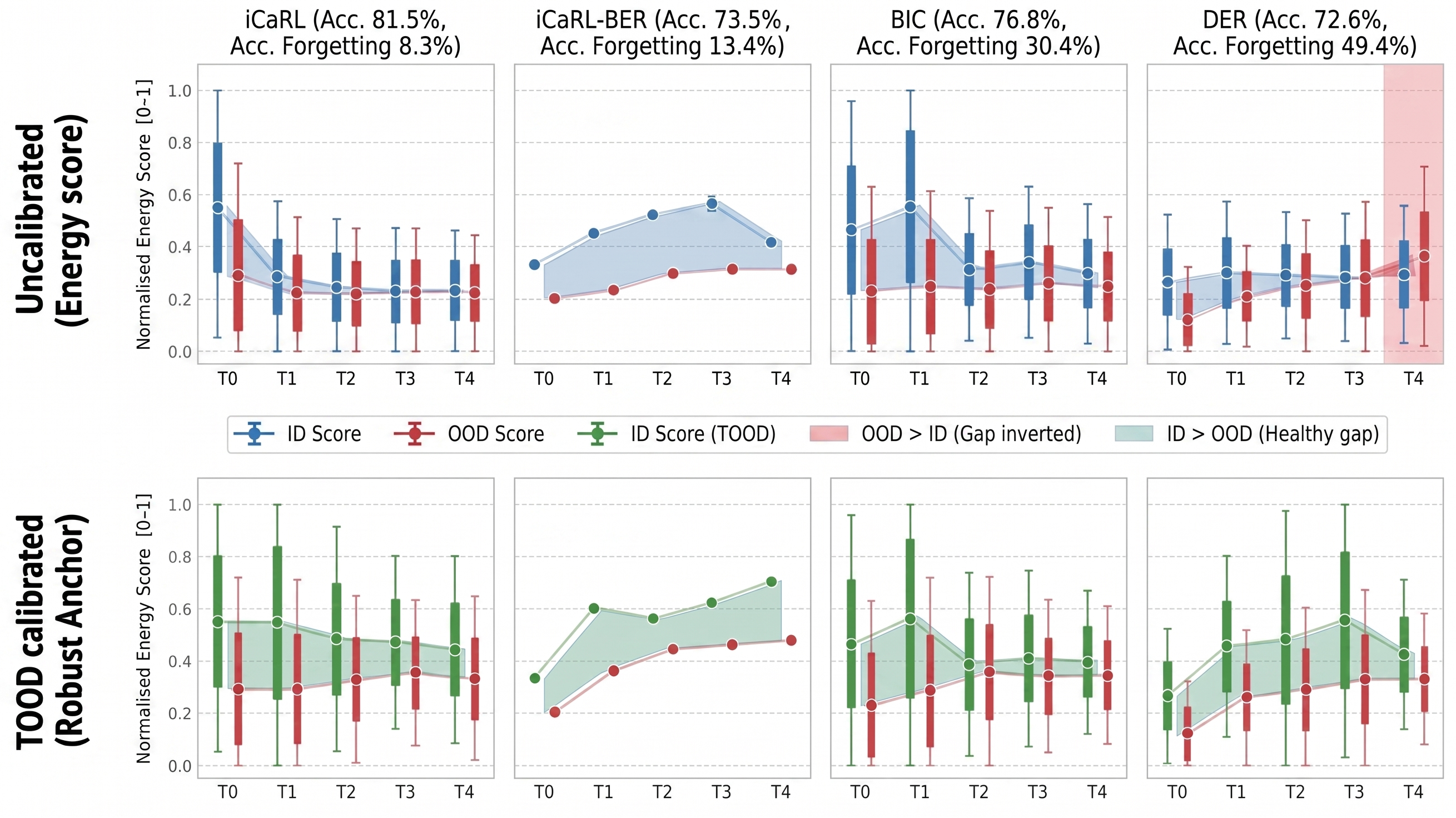}
    \caption{%
        \textbf{TOOD restores \ac{ID}/\ac{OOD} separability across \ac{CL} methods.}
        Each column corresponds to a different \ac{CL} method.
        \emph{Top row:} Uncalibrated Task~0 \ac{ID} energy scores (blue) converge toward \ac{OOD} scores (red) as new tasks are learned, eroding the detection margin.
        \emph{Bottom row:} After \ac{TOOD}'s per-task normalization, calibrated \ac{ID} scores (green) maintain a clear separation from \ac{OOD} scores throughout the task stream.
        Boxes span the interquartile range; whiskers mark the 5th--95th percentile.
        Shaded red regions indicate task steps where the \ac{ID} and \ac{OOD} medians are inverted.
    }
    \label{fig:whisker}
\end{figure}
\Cref{fig:main_panel} (right) illustrates the confidence gap. The mean \ac{ID} energy for Task 0 drops from approximately 18 at $T_0$ to 6 by $T_4$, converging with the \ac{OOD} scores anchored in the 6–8 range. Consequently, a global threshold $\tau$ established at $T_0$ systematically invalidates \ac{ID} samples from earlier tasks. \Cref{fig:whisker} demonstrates that this trend generalizes across four \ac{CL} methods. In every case, the \ac{ID} energy decays into the \ac{OOD} region, and \ac{TOOD} successfully restores separability. Crucially, because this failure mode occurs in the output scores rather than the latent representations, a post-hoc intervention can substantially recover the detection margin without modifying the underlying feature extractor.

\section{Task-Aware OOD Score Calibration}
\label{sec:method}
This section motivates \ac{TOOD} by showing why scalar post-hoc calibration cannot repair the confidence gap and then introduces TOOD's per-task energy formulation. AUROC depends on the \emph{relative ranking} of scores. Any monotonic transformation applied to all samples preserves rank and cannot change AUROC. Since \ac{ID} and \ac{OOD} samples are indistinguishable at test time, a per-task scalar recalibration applied equally would preserve their ordering. Improving AUROC thus requires a transformation that moves \ac{ID} and \ac{OOD} scores in \emph{different directions}. Per-task energy decomposition achieves this. \ac{ID} samples activate one task's energy strongly while \ac{OOD} samples activate none, so the $\max$ operation selects different task channels for different sample types---a differential transformation by construction. Operating purely on logits is a deliberate design choice. It allows \ac{TOOD} to be training-free and architecture-agnostic, applicable to any \ac{CIL} backbone with an expanding classifier head. See \Cref{app:abl_vit} for details on ablations between ResNet and ViT-B/16 backbones (BatchNorm versus LayerNorm), where \ac{TOOD} raises Avg AUROC by $+3.5$ on iCaRL (CIFAR-10) and by up to $+7.4$ on the longer CIFAR-100 stream.

\subsection{Per-Task Energy Decomposition}
\label{sec:decomp}

Let $\mathbf{h}(x) \in \mathbb{R}^{|\mathcal{C}_\text{seen}|}$ be the logit vector, where $\mathcal{C}_\text{seen} = \bigcup_{t=1}^{N} \mathcal{C}_t$. We use the sign-reversed energy convention~\citep{liu2020energy}, so higher scores indicate \ac{ID}: $E(x) = \log \sum_{c \in \mathcal{C}_\text{seen}} \exp(h_c(x))$. This score drifts as old-class logits shrink. \ac{TOOD} replaces this with $N$ per-task energies:
\begin{equation}
    E_t(x) = \log \sum_{c \,\in\, \mathcal{C}_t}
             \exp\bigl(h_c(x)\bigr),
    \label{eq:task_energy}
\end{equation}
yielding $\mathbf{E}(x) = [E_t]_{t=1}^N \in \mathbb{R}^N$. An \ac{ID} sample from task $T_i$ activates $E_i$ strongly and all other $E_t$ weakly; an \ac{OOD} sample activates no task dominantly. This structure persists even after the global score collapses.

\subsection{Task-Wise Score Normalization}
\label{sec:normalization}

Per-task energies still suffer from the confidence gap. $E_i(x)$ for an old task has lower absolute value than $E_N(x)$ for the current task. \ac{TOOD} normalizes each task's energy using task-wise \ac{ID} calibration statistics. We write $\mathcal{B}_t$ for the calibration set (the replay buffer when available, or a small held-out \ac{ID} split otherwise as in regularization-based methods like LwF~\citep{li2017learning}). Two variants are provided.

\medskip
\noindent\textbf{Mean Shift.}
For each task $T_t$, the buffer mean is $\mu_t =  1 / |\mathcal{B}_t| \sum_{x \in \mathcal{B}_t} E_t(x) $
and the corrected energy is
\begin{equation}
    E_t^{\text{ms}}(x) = E_t(x) + (\mu_\text{ref} - \mu_t),
    \label{eq:mean_shift}
\end{equation}
where $\mu_\text{ref}$ is the calibration-set mean of the most recent task $T_N$. Mean shift corrects location drift but assumes stable spread; when calibration sets are small and distributions heavy-tailed, the mean is sensitive to outliers.

\medskip
\noindent\textbf{Robust Anchor.}
Robust anchor substitutes the median and \ac{MAD}~\citep{huber2011robust}:
\begin{equation}
    E_t^{\text{rob}}(x) =
        \frac{E_t(x) - \tilde{e}_t}{\widehat{\text{MAD}}_t}
        \cdot \widehat{\text{MAD}}_\text{ref} + \tilde{e}_\text{ref},
    \label{eq:robust}
\end{equation}
where $\tilde{e}_t = \text{median}_{x \in \mathcal{B}_t}(E_t(x))$
and
$\widehat{\text{MAD}}_t =
\text{median}_{x \in \mathcal{B}_t}\bigl|E_t(x) - \tilde{e}_t\bigr|$, scaled to normal consistency. The reference $(\tilde{e}_\text{ref}, \widehat{\text{MAD}}_\text{ref})$ comes from $T_N$'s calibration set. We call this $T_N$ the \emph{reference task}: its statistics set only the location and scale (the units) of the normalized scores, not their ordering. Substituting a different reference task (e.g., the oldest task, a max-spread task, or the mean of all task statistics) merely replaces $(\tilde{e}_\text{ref}, \widehat{\mathrm{MAD}}_\text{ref})$, which rescales and shifts \emph{every} normalized channel by the same positive factor and offset. This common affine map commutes with the $\max$ in \Cref{eq:tood}, so it preserves the ranking of scores. Because AUROC depends only on that ranking, it is unchanged (largest deviation $0.00015$ across our full sweep; see \Cref{app:abl_anchor} for details). We keep $T_N$ as the default only because it gives an interpretable, current-scale decision threshold. 
Because the median resists outlier contamination from representation drift, robust anchor is the primary variant in all reported experiments. To test the effect of re-centering, we ablate against the same per-task split but only rescale each task's energy by its own spread, with no common reference point. This actually lowers Avg AUROC by $3.6$ points below uncalibrated energy, while \ac{TOOD} improves it by $3.1$ (\Cref{app:abl_mechanism}). The recentering, not task-awareness on its own, is what drives the improvement.

\subsection{Task-Normalized Energy Score}
\label{sec:tne}

After normalization, \ac{TOOD} computes the final detection score as the
maximum normalized energy across all tasks:
\begin{equation}
    S_{\text{TOOD}}(x) =
        \max_{t \in \{1,\ldots,N\}} E_t^{\text{norm}}(x),
    \label{eq:tood}
\end{equation}
where $E_t^{\text{norm}}$ denotes whichever of
\Cref{eq:mean_shift} or \Cref{eq:robust} is selected.

Importantly, \ac{TOOD} \emph{does not require the task identity of a test sample}. The task index in~\Cref{eq:tood} is not provided externally; instead, all task-wise energies are computed and the final score is obtained by taking the maximum over normalized task channels. \ac{TOOD} only uses the training-time task-to-class partition, i.e., which logits correspond to classes introduced in each incremental task. This partition is already present in standard \ac{CIL}, where the classifier head expands as new classes arrive. Thus, \ac{TOOD} reuses training-time class grouping but does not assume task-incremental inference. It is also robust to a finer class partition: splitting each task's classes into smaller nonempty groups reduces Avg AUROC by only $0.5$ points on average across CIFAR-10 and CIFAR-100, five continual-learning methods, and three seeds (\Cref{app:abl_partition}). An \ac{ID} sample activates one task's energy dominantly; after normalization, that channel is corrected upward and $S_{\text{TOOD}}$ is high. An \ac{OOD} sample activates no channel strongly, so all normalized energies remain mediocre and $S_{\text{TOOD}}$ is low. Because the normalization happens \emph{inside} the $\max$, the effective transformation differs for \ac{ID} and \ac{OOD} samples, allowing \ac{TOOD} to alter AUROC unlike ordinary scalar calibration (see~\Cref{app:ridge} for full score distributions). \Cref{alg:tood} summarizes the procedure. The calibration phase requires one forward pass over $|\mathcal{B}| = \sum_t |\mathcal{B}_t|$ task-wise calibration samples and runs once after each task. Inference adds $O(N)$ $\operatorname{LogSumExp}$ operations per sample. Both phases are training-free.

\medskip
\noindent\textbf{Optional heuristic margin term.}
For \ac{ID} samples the gap between the top two normalized energies is often large; for \ac{OOD} samples it is usually smaller. A simple heuristic is therefore to add a margin term:
\begin{equation}
    S_{\lambda}(x) = E_{(1)}^{\text{norm}}(x)
                   + \lambda \bigl(
                       E_{(1)}^{\text{norm}}(x)
                     - E_{(2)}^{\text{norm}}(x)
                     \bigr),
    \label{eq:margin}
\end{equation}
where $E_{(1)}^{\text{norm}} \geq E_{(2)}^{\text{norm}}$ are the two largest normalized energies. When $\lambda = 0$, this reduces to~\Cref{eq:tood}. For $\lambda > 0$, \ac{ID} samples with broad margins are pushed to higher scores. Because this term is heuristic rather than theory-derived, we treat it as optional and analyze its sensitivity in~\Cref{app:ablation_margin}; all main experiments use a fixed $\lambda = 0.5$. A sweep shows methods with a large confidence gap (iCaRL, DER) peak near $\lambda = 0.5$ while methods with native output-head correction (BiC) prefer $\lambda = 0$, so $0.5$ is a robust default that requires no per-method tuning.

\begin{algorithm}[!ht]
\caption{TOOD: Task-Aware OOD Score Calibration}
\label{alg:tood}
\begin{algorithmic}[1]
\Require Frozen model $f_\theta$, task-wise calibration sets $\{\mathcal{B}_t\}_{t=1}^N$,
         variant $v \in \{\text{ms},\, \text{rob}\}$, margin $\lambda \ge 0$
\State $R \gets \emptyset$
\For{$t = 1$ to $N$} \Comment{Calibration phase}
    \State Collect logits $\{\mathbf{h}(x)\}_{x \in \mathcal{B}_t}$ using $f_\theta$
    \State Compute task energies $\{E_t(x)\}_{x \in \mathcal{B}_t}$ via \Cref{eq:task_energy}
    \If{$v = \text{ms}$}
        \State $R[t] \gets \mu_t$ \Comment{Calibration-set mean}
    \ElsIf{$v = \text{rob}$}
        \State $R[t] \gets (\tilde{e}_t, \widehat{\text{MAD}}_t)$ \Comment{Calibration-set median and \ac{MAD}}
    \EndIf
\EndFor
\State $R_{\text{ref}} \gets R[N]$ \Comment{Reference statistics from the most recent task}
\For{each test sample $x$} \Comment{Inference phase}
    \For{$t = 1$ to $N$}
        \State Compute $E_t(x)$ via \Cref{eq:task_energy}
        \State Compute $E_t^{\text{norm}}(x)$ using $R[t]$ and $R_{\text{ref}}$ via \Cref{eq:mean_shift} or \Cref{eq:robust}
    \EndFor
    \State Let $E_{(1)}^{\text{norm}}(x)$ and $E_{(2)}^{\text{norm}}(x)$ be the largest and second-largest normalized energies
    \State $S_{\text{TOOD}}(x) \gets E_{(1)}^{\text{norm}}(x) + \lambda \big(E_{(1)}^{\text{norm}}(x) - E_{(2)}^{\text{norm}}(x)\big)$
\EndFor
\State \Return $\{S_{\text{TOOD}}(x)\}_{x}$ \Comment{OOD scores for all test samples}
\end{algorithmic}
\end{algorithm}

\section{Results}
\label{sec:experiments}

\ac{CL} encompasses three canonical settings~\citep{van2019three}: \emph{task-incremental learning} (TIL), where a task identifier is available at test time; \emph{domain-incremental learning} (DIL), where the input distribution shifts but the label space remains fixed; and \emph{class-incremental} (CIL), where new classes arrive sequentially and no task identifier is provided at inference. We focus on CIL because it maximally stresses \ac{OOD} detection: each new task both expands the classifier head and redefines the semantic boundary between \ac{ID} and \ac{OOD} data, producing the strongest confidence gap and the most acute manifold crowding. TIL, by contrast, sidesteps output-head interference through task-conditioned prediction, while DIL keeps the label space fixed and thus introduces no head expansion.

Across datasets, \ac{TOOD} recovers \ac{OOD} performance lost in continual learning, especially under severe logit-scale drift; extension to  task- and domain-incremental CL remains future work.

\subsection{Experimental Setup}
\label{sec:setup}

\paragraph{Datasets.}

For training the \ac{CL} systems we primarily use CIFAR-10~\citep{krizhevsky2009learning} ($N\!=\!5$ tasks, 2 classes each) and CIFAR-100~\citep{krizhevsky2009learning} ($N\!=\!10$ tasks, 10 classes each). We also use ImageNet-1K~\citep{deng2009imagenet} with $ N\!=\!100 $ tasks, 10 classes each. For evaluating \ac{OOD} detection, we use the \ac{OOD} datasets from OpenOOD~\citep{zhang2023openood}, and we follow the Near-Far OOD labeling from OpenOOD with respect to CIFAR-10, CIFAR-100, and ImageNet-1k. For CIFAR-10/100, near-\ac{OOD} datasets are CIFAR-100/TinyImageNet, and far-\ac{OOD} datasets are MNIST~\citep{yann2010mnist}, SVHN~\citep{netzer2011reading}, Textures~\citep{cimpoi2014describing}, and Places365~\citep{zhou2017places}. 
For ImageNet-1k, near-\ac{OOD} datasets are SSB-hard/NINCO~\citep{bitterwolf2023ninco}, and far-\ac{OOD} are iNaturalist~\citep{huang2021mos}, Textures~\citep{kylberg2011kylberg}, and OpenImage-O~\citep{haoqi2022vim}. To generate replay buffers $\mathcal{B}_t$ for TOOD calibration, we use a randomly generated class-balanced buffer~\citep{rebuffi2017icarl}, 

\paragraph{CL Methods and Curricula.}
We test TOOD on five representative \ac{CL} methods spanning three paradigms: \textit{distillation / regularization}: LwF~\citep{li2017learning}; \textit{replay}: iCaRL~\citep{rebuffi2017icarl}, DER~\citep{buzzega2020dark}; \textit{output correction}: BiC~\citep{wu2019large}, WA~\citep{zhao2019maintaining}. All methods are implemented via the Avalanche library~\citep{lomonacoavalanche} using the hyper-parameters recommended by the authors. We train each CL method following the class-label curriculum established by prior CIL benchmarks~\citep{rebuffi2017icarl,wu2019large,zhao2019maintaining}. These curricula assign classes in label-index order with equal cardinality per task, with the goal that each task introduces enough classes  to produce a meaningful shift in the output head while keeping per-task training cost manageable. CIFAR-10 is split into $N\!=\!5$ tasks of 2 classes, CIFAR-100 into $N\!=\!10$ tasks of 10 classes, and ImageNet-1K into $N\!=\!100$ tasks of 10 classes. All models are trained for 170 epochs (SGD, momentum $0.9$, batch size $128$, Multi-Step LR from $0.1$ with $\gamma\!=\!0.1$ at milestones $[60, 100, 140]$). CIFAR-10/100 datasets use a ResNet-32~\citep{he2016deep} backbone with a single expanding linear head, while ImageNet-1k uses a ResNet-18.

\paragraph{OOD Detectors.} 
We compare TOOD against eight post-hoc OOD detection baselines, spanning three detection paradigms: \textit{output-based}: MSP~\citep{hendrycks2017baseline}, Energy~\citep{liu2020energy}; Dice~\citep{sun2021dice}, ADASCALE~\citep{regmi2025adascale};  \textit{activation shaping}: ASH~\citep{djurisic2023extremely}; \textit{feature-based}: ViM~\citep{haoqi2022vim}, MDS~\citep{sun2022out}, NNGuide~\citep{park2023nearest}. All methods are implemented via the OpenOOD benchmark~\citep{zhang2023openood}. The above baselines are compared against both TOOD variants, Energy + Mean Shift and Energy + Robust Anchor. We also compare TOOD against bi-directional energy regularization (BER) from OpenCIL~\citep{miao2024opencil}, a finetuning method for improving OOD detection in CIL.

\paragraph{Code.} Code is available at \url{https://github.com/mostafaelaraby/tood-continual-ood}.

\subsection{Main Results}

\Cref{tab:main_results} summarizes the main results of our comparisons between \ac{TOOD} and other \ac{OOD} detectors. With 3-seed mean-plus-std estimates, at least one TOOD variant ranks first or second in eight of ten dataset and CL method combinations, ranked by Avg AUROC. Many other methods lack consistency, for example ASH fluctuates across CL methods (e.g., $65.9$ on iCaRL but $57.7$ on DER). Some feature-based methods, such as MDS, which are typically strong in non-CL settings struggle relative to TOOD (e.g. $58.4$ vs. $66.5$ on iCaRL and CIFAR-10), corroborating our manifold crowding diagnosis.

\Cref{tab:main_results} also provides more evidence for the importance of the confidence gap. BiC~\citep{wu2019large}, which applies explicit output-head bias correction, reaches $67.2$ AUROC on CIFAR-10, higher than DER ($61.2$) despite lower classification accuracy. iCaRL and DER stabilize logit values for replayed samples but impose no constraint on absolute scale, showing correspondingly lower AUROC. Moreover, \ac{TOOD} delivers its largest improvements when logit-scale drift is severe. On CIFAR-10, Mean Shift improves DER by $+8.1$ AUROC ($61.2 \to 69.3$) and BiC by $+4.4$ ($67.2 \to 71.6$). Conversely, for methods that natively stabilize output scales (e.g., WA), the confidence gap is narrower, yielding marginal or statistically insignificant changes (e.g., WA on CIFAR-100 shifts $71.9 \to 71.7$). Thus, \ac{TOOD} acts as a targeted first-order repair for score drift rather than a universal enhancement across all \ac{CL} methods. 
\begin{table*}[t]
\centering\footnotesize
\setlength{\tabcolsep}{2pt}
\caption{%
    Main results across 2 benchmarks.
    Columns are CL methods; rows are OOD detectors.
    We report average CIL accuracy, \textbf{Avg AUROC}~(\Cref{eq:avg_auroc}) ($\uparrow$), and \textbf{Avg FPR@95} ($\downarrow$), with AUROC and FPR@95 averaged over Near- and Far-OOD.
    Values are mean$\pm$std over 3 seeds.
    Bold marks every entry within one std of the best per metric subcolumn.}
\label{tab:main_results}
\begin{tabular}{l cc cc cc cc cc}
\toprule
\multicolumn{11}{c}{\textbf{(a) CIFAR-10 ($N = 5$)}} \\
\midrule
\textbf{OOD Method} & \multicolumn{2}{c}{iCaRL} & \multicolumn{2}{c}{BiC} & \multicolumn{2}{c}{DER} & \multicolumn{2}{c}{WA} & \multicolumn{2}{c}{LwF} \\
 & AUC $\uparrow$ & FPR $\downarrow$ & AUC $\uparrow$ & FPR $\downarrow$ & AUC $\uparrow$ & FPR $\downarrow$ & AUC $\uparrow$ & FPR $\downarrow$ & AUC $\uparrow$ & FPR $\downarrow$ \\
\midrule
Avg CIL Acc. ($\uparrow$) & \multicolumn{2}{c}{80.9\,{\scriptsize$\pm$0.3}} & \multicolumn{2}{c}{77.5\,{\scriptsize$\pm$0.2}} & \multicolumn{2}{c}{72.7\,{\scriptsize$\pm$0.2}} & \multicolumn{2}{c}{80.2\,{\scriptsize$\pm$0.9}} & \multicolumn{2}{c}{67.9\,{\scriptsize$\pm$0.1}} \\
\midrule
MSP & 63.0\,{\scriptsize$\pm$1.4} & 82.4\,{\scriptsize$\pm$0.8} & 68.3\,{\scriptsize$\pm$0.6} & 77.0\,{\scriptsize$\pm$3.6} & \textbf{69.0}\,{\scriptsize$\pm$0.6} & 71.1\,{\scriptsize$\pm$1.0} & 74.2\,{\scriptsize$\pm$2.1} & 67.6\,{\scriptsize$\pm$8.1} & 66.3\,{\scriptsize$\pm$0.6} & 72.3\,{\scriptsize$\pm$0.4} \\
Ash & \textbf{65.9}\,{\scriptsize$\pm$1.7} & \textbf{71.8}\,{\scriptsize$\pm$1.1} & 61.9\,{\scriptsize$\pm$2.1} & 71.5\,{\scriptsize$\pm$2.1} & 57.7\,{\scriptsize$\pm$1.0} & 79.3\,{\scriptsize$\pm$0.8} & 62.2\,{\scriptsize$\pm$1.7} & 73.5\,{\scriptsize$\pm$0.3} & 64.7\,{\scriptsize$\pm$1.2} & 74.8\,{\scriptsize$\pm$0.7} \\
VIM & 63.9\,{\scriptsize$\pm$1.2} & 81.7\,{\scriptsize$\pm$1.5} & 65.7\,{\scriptsize$\pm$1.5} & 73.5\,{\scriptsize$\pm$1.1} & 61.1\,{\scriptsize$\pm$0.3} & 77.8\,{\scriptsize$\pm$0.9} & \textbf{77.9}\,{\scriptsize$\pm$2.0} & \textbf{61.4}\,{\scriptsize$\pm$3.5} & \textbf{72.9}\,{\scriptsize$\pm$1.4} & 70.1\,{\scriptsize$\pm$1.5} \\
Dice & 61.5\,{\scriptsize$\pm$1.5} & \textbf{72.7}\,{\scriptsize$\pm$1.0} & 54.9\,{\scriptsize$\pm$0.7} & 74.9\,{\scriptsize$\pm$1.3} & 53.7\,{\scriptsize$\pm$1.4} & 79.5\,{\scriptsize$\pm$1.0} & 59.4\,{\scriptsize$\pm$0.8} & 72.5\,{\scriptsize$\pm$0.3} & 62.9\,{\scriptsize$\pm$2.9} & 75.3\,{\scriptsize$\pm$0.9} \\
ADASCALE & 58.0\,{\scriptsize$\pm$2.1} & 85.9\,{\scriptsize$\pm$1.7} & 58.5\,{\scriptsize$\pm$2.3} & 79.7\,{\scriptsize$\pm$3.3} & 52.6\,{\scriptsize$\pm$1.2} & 85.1\,{\scriptsize$\pm$0.7} & 62.6\,{\scriptsize$\pm$0.6} & 81.5\,{\scriptsize$\pm$1.2} & 63.9\,{\scriptsize$\pm$1.5} & 81.2\,{\scriptsize$\pm$2.6} \\
MDS & 58.4\,{\scriptsize$\pm$0.1} & 85.5\,{\scriptsize$\pm$1.5} & 61.1\,{\scriptsize$\pm$1.1} & 84.3\,{\scriptsize$\pm$0.9} & 59.3\,{\scriptsize$\pm$3.0} & 86.6\,{\scriptsize$\pm$2.9} & 54.2\,{\scriptsize$\pm$1.4} & 89.8\,{\scriptsize$\pm$2.8} & 53.7\,{\scriptsize$\pm$1.8} & 90.0\,{\scriptsize$\pm$2.6} \\
NNGuide & 50.1\,{\scriptsize$\pm$1.3} & 85.4\,{\scriptsize$\pm$1.2} & 48.8\,{\scriptsize$\pm$1.2} & 79.0\,{\scriptsize$\pm$1.0} & 43.4\,{\scriptsize$\pm$0.6} & 86.9\,{\scriptsize$\pm$0.3} & 58.4\,{\scriptsize$\pm$1.3} & 74.0\,{\scriptsize$\pm$2.3} & 55.8\,{\scriptsize$\pm$2.6} & 79.1\,{\scriptsize$\pm$1.1} \\
Energy (Uncalib.) & \textbf{65.9}\,{\scriptsize$\pm$1.5} & 77.6\,{\scriptsize$\pm$0.9} & 67.2\,{\scriptsize$\pm$1.0} & 69.1\,{\scriptsize$\pm$1.1} & 61.2\,{\scriptsize$\pm$0.7} & 74.1\,{\scriptsize$\pm$0.4} & \textbf{78.1}\,{\scriptsize$\pm$2.3} & \textbf{59.5}\,{\scriptsize$\pm$4.5} & 72.2\,{\scriptsize$\pm$0.5} & 70.4\,{\scriptsize$\pm$0.8} \\
\midrule
Energy + Mean Shift & \textbf{66.5}\,{\scriptsize$\pm$1.5} & 75.4\,{\scriptsize$\pm$1.0} & \textbf{71.6}\,{\scriptsize$\pm$0.2} & \textbf{66.9}\,{\scriptsize$\pm$0.3} & \textbf{69.3}\,{\scriptsize$\pm$0.9} & \textbf{68.3}\,{\scriptsize$\pm$0.7} & \textbf{78.3}\,{\scriptsize$\pm$2.1} & \textbf{59.3}\,{\scriptsize$\pm$4.6} & \textbf{73.2}\,{\scriptsize$\pm$0.3} & \textbf{68.6}\,{\scriptsize$\pm$0.5} \\
Energy + Robust Anchor & \textbf{66.4}\,{\scriptsize$\pm$1.4} & 75.6\,{\scriptsize$\pm$0.9} & 70.6\,{\scriptsize$\pm$0.3} & 69.0\,{\scriptsize$\pm$0.8} & \textbf{68.6}\,{\scriptsize$\pm$0.9} & \textbf{68.9}\,{\scriptsize$\pm$0.8} & \textbf{78.2}\,{\scriptsize$\pm$2.4} & \textbf{59.8}\,{\scriptsize$\pm$4.9} & \textbf{73.0}\,{\scriptsize$\pm$0.7} & \textbf{68.4}\,{\scriptsize$\pm$0.8} \\
\bottomrule
\end{tabular}

\vspace{2mm}

\begin{tabular}{l cc cc cc cc cc}
\toprule
\multicolumn{11}{c}{\textbf{(b) CIFAR-100 ($N = 10$)}} \\
\midrule
\textbf{OOD Method} & \multicolumn{2}{c}{iCaRL} & \multicolumn{2}{c}{BiC} & \multicolumn{2}{c}{DER} & \multicolumn{2}{c}{WA} & \multicolumn{2}{c}{LwF} \\
 & AUC $\uparrow$ & FPR $\downarrow$ & AUC $\uparrow$ & FPR $\downarrow$ & AUC $\uparrow$ & FPR $\downarrow$ & AUC $\uparrow$ & FPR $\downarrow$ & AUC $\uparrow$ & FPR $\downarrow$ \\
\midrule
Avg CIL Acc. ($\uparrow$) & \multicolumn{2}{c}{53.0\,{\scriptsize$\pm$0.1}} & \multicolumn{2}{c}{62.3\,{\scriptsize$\pm$0.1}} & \multicolumn{2}{c}{55.1\,{\scriptsize$\pm$0.4}} & \multicolumn{2}{c}{63.8\,{\scriptsize$\pm$0.4}} & \multicolumn{2}{c}{50.7\,{\scriptsize$\pm$0.4}} \\
\midrule
MSP & \textbf{61.5}\,{\scriptsize$\pm$1.0} & \textbf{81.0}\,{\scriptsize$\pm$1.0} & 64.8\,{\scriptsize$\pm$0.4} & 79.1\,{\scriptsize$\pm$0.3} & 65.9\,{\scriptsize$\pm$0.2} & \textbf{75.6}\,{\scriptsize$\pm$0.6} & 69.8\,{\scriptsize$\pm$0.2} & 71.7\,{\scriptsize$\pm$0.1} & 64.4\,{\scriptsize$\pm$1.0} & 80.8\,{\scriptsize$\pm$1.3} \\
Ash & 59.0\,{\scriptsize$\pm$2.7} & 85.0\,{\scriptsize$\pm$1.3} & 51.1\,{\scriptsize$\pm$1.0} & 87.7\,{\scriptsize$\pm$0.7} & 46.4\,{\scriptsize$\pm$2.0} & 91.0\,{\scriptsize$\pm$1.2} & 53.5\,{\scriptsize$\pm$1.8} & 86.9\,{\scriptsize$\pm$1.0} & 64.9\,{\scriptsize$\pm$0.8} & \textbf{78.0}\,{\scriptsize$\pm$1.6} \\
VIM & 57.9\,{\scriptsize$\pm$0.8} & 83.8\,{\scriptsize$\pm$0.7} & 60.1\,{\scriptsize$\pm$0.3} & 81.0\,{\scriptsize$\pm$0.7} & 61.4\,{\scriptsize$\pm$0.4} & 80.2\,{\scriptsize$\pm$0.5} & 70.3\,{\scriptsize$\pm$0.2} & 71.1\,{\scriptsize$\pm$0.2} & \textbf{67.3}\,{\scriptsize$\pm$1.9} & \textbf{76.7}\,{\scriptsize$\pm$2.5} \\
Dice & 60.4\,{\scriptsize$\pm$0.7} & 83.4\,{\scriptsize$\pm$1.1} & 52.1\,{\scriptsize$\pm$0.4} & 86.4\,{\scriptsize$\pm$0.6} & 50.2\,{\scriptsize$\pm$1.7} & 86.6\,{\scriptsize$\pm$1.1} & 55.1\,{\scriptsize$\pm$2.6} & 85.3\,{\scriptsize$\pm$2.1} & 60.2\,{\scriptsize$\pm$1.4} & 80.8\,{\scriptsize$\pm$2.0} \\
ADASCALE & 58.3\,{\scriptsize$\pm$0.9} & 87.4\,{\scriptsize$\pm$1.9} & 57.2\,{\scriptsize$\pm$0.5} & 88.5\,{\scriptsize$\pm$0.5} & 57.6\,{\scriptsize$\pm$2.1} & 88.7\,{\scriptsize$\pm$1.4} & 67.8\,{\scriptsize$\pm$1.2} & 80.8\,{\scriptsize$\pm$1.6} & 64.6\,{\scriptsize$\pm$1.3} & 84.6\,{\scriptsize$\pm$1.4} \\
MDS & 57.7\,{\scriptsize$\pm$1.8} & 84.7\,{\scriptsize$\pm$1.4} & 62.3\,{\scriptsize$\pm$0.4} & 79.5\,{\scriptsize$\pm$0.5} & 59.1\,{\scriptsize$\pm$1.7} & 83.2\,{\scriptsize$\pm$1.4} & 58.5\,{\scriptsize$\pm$0.9} & 82.5\,{\scriptsize$\pm$1.1} & 53.9\,{\scriptsize$\pm$0.4} & 87.9\,{\scriptsize$\pm$0.5} \\
NNGuide & 58.1\,{\scriptsize$\pm$1.6} & \textbf{81.8}\,{\scriptsize$\pm$0.9} & 52.0\,{\scriptsize$\pm$1.6} & 85.4\,{\scriptsize$\pm$0.7} & 55.3\,{\scriptsize$\pm$0.3} & 83.8\,{\scriptsize$\pm$0.4} & 53.5\,{\scriptsize$\pm$1.4} & 83.5\,{\scriptsize$\pm$0.5} & 52.9\,{\scriptsize$\pm$0.5} & 84.3\,{\scriptsize$\pm$0.3} \\
Energy (Uncalib.) & 58.1\,{\scriptsize$\pm$1.3} & 83.0\,{\scriptsize$\pm$0.9} & 65.6\,{\scriptsize$\pm$0.5} & 78.2\,{\scriptsize$\pm$0.4} & 64.6\,{\scriptsize$\pm$1.2} & 77.3\,{\scriptsize$\pm$1.2} & \textbf{71.9}\,{\scriptsize$\pm$0.1} & \textbf{68.3}\,{\scriptsize$\pm$0.5} & \textbf{67.0}\,{\scriptsize$\pm$1.6} & \textbf{76.3}\,{\scriptsize$\pm$2.0} \\
\midrule
Energy + Mean Shift & 58.4\,{\scriptsize$\pm$1.0} & 82.5\,{\scriptsize$\pm$0.7} & \textbf{68.4}\,{\scriptsize$\pm$1.3} & \textbf{74.7}\,{\scriptsize$\pm$1.1} & \textbf{67.1}\,{\scriptsize$\pm$0.5} & \textbf{75.8}\,{\scriptsize$\pm$1.2} & 71.7\,{\scriptsize$\pm$0.2} & 68.8\,{\scriptsize$\pm$0.9} & \textbf{66.7}\,{\scriptsize$\pm$1.7} & \textbf{77.4}\,{\scriptsize$\pm$2.1} \\
Energy + Robust Anchor & 58.9\,{\scriptsize$\pm$1.0} & 82.4\,{\scriptsize$\pm$0.5} & \textbf{68.8}\,{\scriptsize$\pm$1.6} & \textbf{74.7}\,{\scriptsize$\pm$1.8} & 66.1\,{\scriptsize$\pm$0.9} & \textbf{76.1}\,{\scriptsize$\pm$1.1} & 71.5\,{\scriptsize$\pm$0.1} & 69.4\,{\scriptsize$\pm$0.9} & \textbf{66.3}\,{\scriptsize$\pm$1.6} & \textbf{77.8}\,{\scriptsize$\pm$1.8} \\
\bottomrule
\end{tabular}
\end{table*}

\Cref{tab:ber_comparison} summarizes our results comparing TOOD to the most related existing method, BER; the post-hoc rows (Energy, TOOD) are 3-seed means while the 10-epoch BER rows are single-seed. While the two methods are in theory compatible, combining them does not consistently outperform the better of the two alone, and in five of eight metric columns a post-hoc TOOD variant outperforms all BER rows. Moreover, BER exhibits instability across scales: while it improves iCaRL on CIFAR-10, it degrades it on CIFAR-100 ($58.1 \to 57.7$). Furthermore, TOOD requires no retraining, making it a simpler and cheaper option in most cases. 
\begin{table}[ht]
\centering\small
\setlength{\tabcolsep}{4pt}
\caption{%
    Post-hoc Calibration (TOOD) versus 10-Epoch fine-tuning (BER).
    We report \textbf{Avg AUROC}~(\Cref{eq:avg_auroc}) ($\uparrow$) and \textbf{Avg FPR@95} ($\downarrow$), each averaged over Near- and Far-OOD. Bold indicates 1st place per metric subcolumn. The post-hoc rows (Energy, TOOD) are means over 3 seeds and match \Cref{tab:main_results}; the 10-epoch BER and BER\,+\,TOOD rows are reported from a single seed.
}
\label{tab:ber_comparison}
\begin{tabular}{l cc cc | cc cc}
\toprule
 & \multicolumn{4}{c|}{\textbf{CIFAR-10 ($N = 5$)}} & \multicolumn{4}{c}{\textbf{CIFAR-100 ($N = 10$)}} \\
\cmidrule(lr){2-5}\cmidrule(lr){6-9}
\textbf{Method} & \multicolumn{2}{c}{iCaRL} & \multicolumn{2}{c}{BiC} & \multicolumn{2}{c}{iCaRL} & \multicolumn{2}{c}{BiC} \\
 & AUC $\uparrow$ & FPR $\downarrow$ & AUC $\uparrow$ & FPR $\downarrow$ & AUC $\uparrow$ & FPR $\downarrow$ & AUC $\uparrow$ & FPR $\downarrow$ \\
\midrule
\multicolumn{9}{l}{No Retraining (Zero-Epoch)} \\
Energy (Uncalib.) & 65.9 & 77.6 & 67.2 & 69.1 & 58.1 & 83.0 & 65.6 & 78.2 \\
TOOD (Mean Shift) & 66.5 & 75.4 & \textbf{71.6} & \textbf{66.9} & 58.4 & 82.5 & 68.4 & 74.7 \\
TOOD (Robust Anchor) & 66.4 & 75.6 & 70.6 & 69.0 & \textbf{58.9} & \textbf{82.4} & \textbf{68.8} & 74.7 \\
\midrule
\multicolumn{9}{l}{10-Epoch Retraining Required} \\
BER (Uncalib. Energy) & 73.5 & \textbf{67.1} & 69.5 & 69.1 & 57.7 & 84.0 & 68.5 & \textbf{74.1} \\
BER + TOOD (Mean Shift) & \textbf{74.3} & 68.1 & 68.0 & 69.6 & 58.4 & 84.4 & 66.9 & 74.9 \\
BER + TOOD (Robust Anchor) & 73.6 & 70.6 & 66.1 & 77.6 & 55.9 & 86.3 & 68.7 & 75.7 \\
\bottomrule
\end{tabular}
\end{table}

\subsection{Ablation Studies}

\begin{figure*}[!ht]
    \centering
    \includegraphics[width=0.9\linewidth]{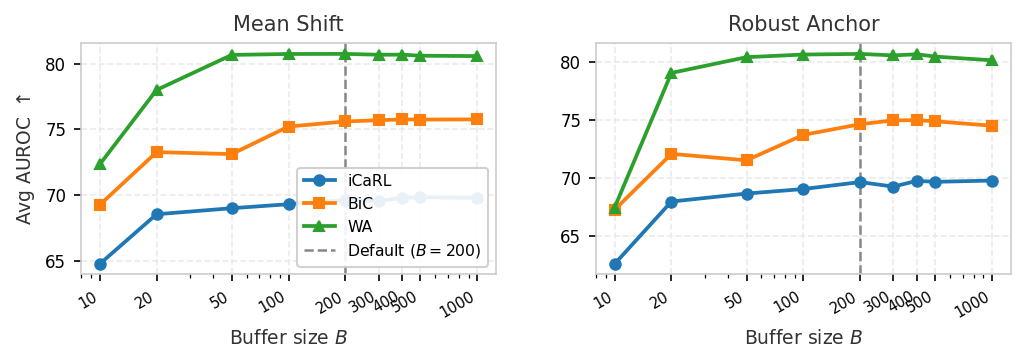}
    \caption{%
        \textbf{Effect of buffer size $B$ (CIFAR-10, $N=5$ tasks).}
        Average \ac{AUROC} ($\uparrow$, top) and \ac{OOD} deterioration ($\downarrow$, bottom) saturate quickly. The default $B = 200$ (dashed line) is well within the stable performance plateau, confirming the robustness of our results to buffer constraints.
    }
    \label{fig:buffer_ablation}
\end{figure*}

\textbf{Replay Buffer Size.}
\ac{TOOD} relies on a replay buffer to estimate per-task statistics ($\mu_t$, $\tilde{e}_t$, $\widehat{\text{MAD}}_t$). In our experiments we use a buffer size of $B = 200$ (CIFAR-10) and $B = 700$ (CIFAR-100) to contain samples across all tasks $T_0 \ldots T_N$. \Cref{fig:buffer_ablation} shows TOOD performance over $B$. Performance is unstable for $B < 20$ but stabilizes between $B=50$ and $B=100$, with negligible gains thereafter. Thus, \ac{TOOD} requires only a few dozen samples per task; our default values $200$ and $700$ are well above this threshold.
We also tested iCaRL-style herding instead of random class-balanced calibration samples. The average change in Avg AUROC is $-0.02$ points over the five Table~\ref{tab:main_results} \ac{CL} methods and two CIFAR datasets, with no consistent gain, so random class-balanced calibration is sufficient for the median/\ac{MAD} statistics used by \ac{TOOD}.

\section{Conclusion}
\label{sec:conclusion}
In this paper we show that standard CL evaluations focused on classification forgetting and feature drift do not predict a CL system's ability to reject OOD inputs, and that this capability degrades substantially over the task stream. This mismatch indicates OOD forgetting happens via alternate mechanisms, of which we identify two: manifold crowding and the confidence gap, which describe i) how newly learned tasks consume feature-space margin and weaken distance-based separation and ii) the tendency for old-task logits to shrink in absolute scale relative to newer tasks, respectively. Motivated by these insights, we propose TOOD, a post-hoc, training-free algorithm that decomposes logits into per-task energy scores and re-calibrates them using task-wise \ac{ID} statistics. We show that TOOD compares favorably to many OOD detection methods on a variety of CL datasets and CL training methods. Of course, TOOD does not prevent OOD forgetting outright. It relies on the training-time task-to-class grouping that \ac{CIL} models already maintain, rather than any test-time task identity, together with a modest amount of \ac{ID} calibration data, so its efficacy likely drops when that grouping is ambiguous or calibration data is scarce; the margin term is optional and can be disabled at no cost. Moreover, while it effectively re-calibrates energy-based OOD detectors, it cannot fix feature-space degradation associated with manifold crowding. In the future, we plan to further analyze manifold crowding and develop OOD detection methods for CL systems that mitigate this problem directly, for example through representation regularization or margin-preserving replay objectives.

\bibliography{references}
\bibliographystyle{collas2026_conference}

\appendix
\crefalias{section}{appendix}
\crefalias{subsection}{appendix}
\crefalias{subsubsection}{appendix}
\clearpage
\section{Experimental and Implementation Details}
\label{app:implementation}

For completeness, all experiments use the datasets, task curricula, continual-learning methods, and OOD detectors specified in \Cref{sec:setup}. CIFAR-10 and CIFAR-100 use ResNet-32 over 5 and 10 tasks, respectively, while ImageNet-1K uses ResNet-18 over 100 tasks. Models are trained for 170 epochs with SGD (momentum $0.9$, batch size $128$, initial learning rate $0.1$, and a factor-$0.1$ decay at epochs 60, 100, and 140). \ac{TOOD} is applied post hoc using random class-balanced calibration sets of $B=200$ for CIFAR-10 and $B=700$ for CIFAR-100. Main CIFAR results use three seeds; ImageNet-1K, the controlled toy experiment, and the ViT experiments are single-seed scalability or diagnostic studies.

\section{Confidence-Gap and Manifold-Crowding Diagnostics}
\label{app:diagnostics}

\begin{figure*}[!ht]
    \centering
    \begin{subfigure}{\linewidth}
        \centering
        \includegraphics[width=\linewidth]{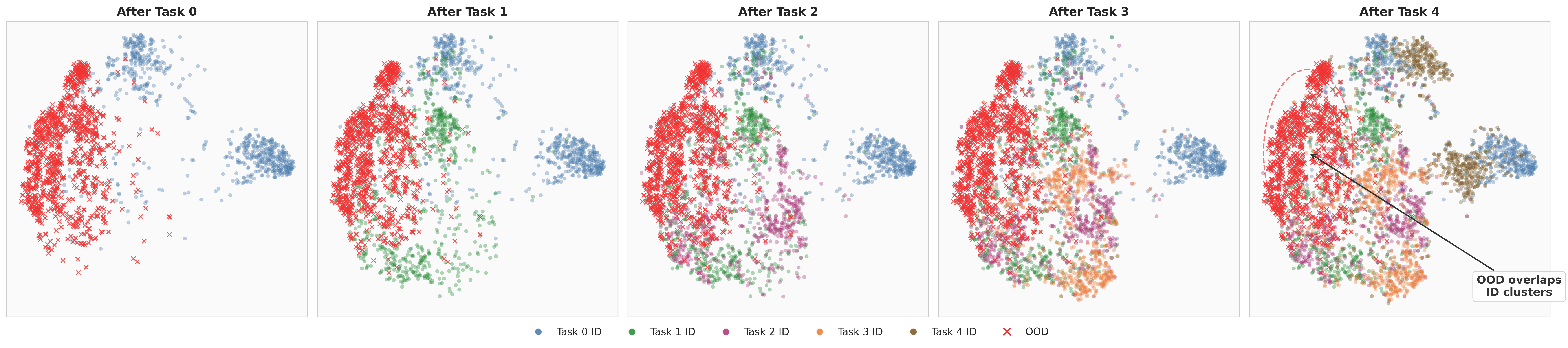}
        \caption{%
            \textbf{OOD samples become surrounded by ID clusters as tasks accumulate.}
            t-SNE projections after each task step (iCaRL, CIFAR-10, \ac{OOD}: SVHN).
            \ac{ID} samples are filled circles (each color a different task);
            \ac{OOD} samples are red crosses.
            At Task~0, \ac{OOD} samples occupy peripheral space; by Task~4, new task
            embeddings have colonized the surrounding regions, leaving \ac{OOD} samples
            less geometrically distinguishable from \ac{ID} data.
        }
        \label{fig:tsne_crowding}
    \end{subfigure}

    \vspace{1.5em}

    \begin{subfigure}{\linewidth}
        \centering
        \includegraphics[width=\linewidth]{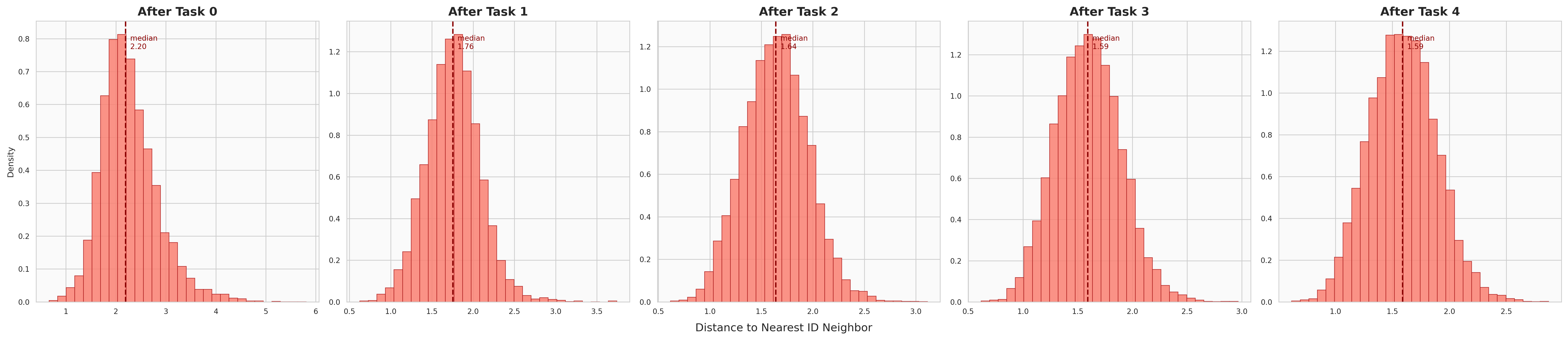}
        \caption{%
            \textbf{The OOD--ID geometric margin shrinks monotonically across tasks.}
            Distribution of Euclidean distances from each \ac{OOD} sample to its nearest
            \ac{ID} neighbor, one curve per task step. The median (dashed line) decreases
            from $2.20$ to $1.59$ (${\sim}28\%$ reduction), with the entire distribution
            shifting leftward by Task~4.
        }
        \label{fig:nn_distance}
    \end{subfigure}

    \caption{%
        \textbf{Manifold Crowding: new task embeddings erode the \ac{OOD} detection
        margin without displacing old representations (iCaRL, CIFAR-10, 5 tasks).}
        Top: the feature space is colonized task by task (t-SNE).
        Bottom: the resulting margin collapse is measured by nearest-neighbor distances.
    }
    \label{fig:manifold_crowding}
\end{figure*} 
\subsection{Manifold crowding}
\label{app:manifold_crowding}

This section provides a detailed characterization of the \textbf{Manifold Crowding} mechanism introduced in~\Cref{sec:manifold_crowding}. While the Confidence Gap operates at the logit level (score compression), Manifold Crowding represents a degradation of the feature-space geometry itself.

\paragraph{Geometric View: t-SNE Projections.}
\Cref{fig:tsne_crowding} visualizes the colonization of the feature space as tasks accumulate. Initially (Task 0), the model maintains a clear separation between \ac{ID} and \ac{OOD} samples. However, as the learning stream progresses, new task embeddings fill the surrounding latent space. By Task 4, \ac{OOD} samples are surrounded by various \ac{ID} clusters, even if the individual clusters remain cohesive. This ``crowding'' explains why distance-based detectors (e.g., MDS) deteriorate: the margin shrinks from the \emph{outside} as new tasks occupy previously empty regions.

\paragraph{Quantitative View: Feature-Space Margin.}
We quantify this erosion in~\Cref{fig:nn_distance} by measuring the Euclidean distance from each \ac{OOD} sample to its nearest \ac{ID} neighbor. The median distance decreases monotonically from $2.20$ to $1.59$ (${\sim}28\%$ reduction). Notably, \ac{CKA} analysis (\Cref{fig:cka}) shows that feature drift remains low, suggesting that the margin loss is driven by the \emph{occupancy} of the space by new classes rather than the movement of old ones.

\subsection{What TOOD does not fix: manifold crowding}
\label{app:abl_crowding}

The confidence gap affects output scores. Manifold crowding instead reduces
the separation between \ac{ID} and \ac{OOD} samples in the learned
representation. Since \ac{TOOD} only recalibrates logits, it cannot restore
this lost geometric margin. We test this by applying the same normalization
separately within each task to nearest prototype distances. This changes Avg
AUROC by $-0.76$ points relative to the original distance detector
(\Cref{fig:mechanism_dissociation}), showing that adjusting the score centers
does not recover geometric separation. Addressing manifold crowding therefore
requires changes to the learned representation, such as preserving margins
through replay or regularization. We leave this direction to future work.

\subsection{Confidence-gap ridge plots}
\label{app:ridge}

\begin{figure*}[!ht]
    \centering
    \includegraphics[width=0.9\linewidth]{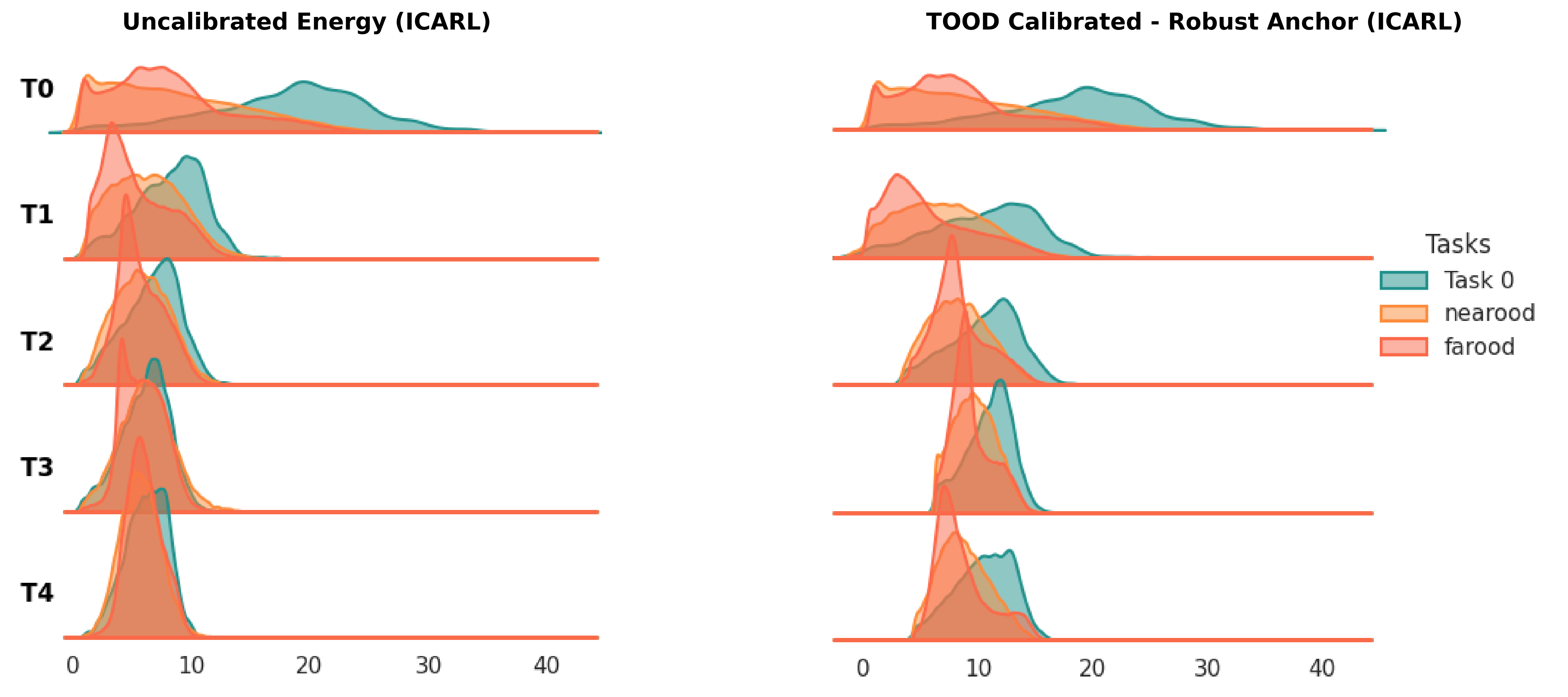}
    \caption{%
        \textbf{TOOD selectively shifts the \ac{ID} score distribution rightward
        while \ac{OOD} scores remain anchored (iCaRL, CIFAR-10, $N=5$ tasks).}
        Each ridge is the energy distribution at one task step.
        \emph{Left (uncalibrated):} the \ac{ID} distribution (blue) shifts progressively
        leftward with each new task, converging with Near-\ac{OOD} (coral) and
        Far-\ac{OOD} (orange) by Task~4, explaining the AUROC collapse.
        \emph{Right (\ac{TOOD}, Robust Anchor):} per-task normalization stabilizes
        the \ac{ID} distribution (blue) at high energy values, while \ac{OOD}
        distributions (coral, orange) remain anchored at low values throughout the stream.
    }
    \label{fig:ridge}
\end{figure*}
\Cref{fig:ridge} provides a distributional view of the confidence gap mechanism introduced in~\Cref{sec:gap}. While the main text quantifies this phenomenon through aggregate metrics (AUROC, $D_{\text{avg}}$), the ridge plots expose the full shape of the score distributions at each task step, making both the failure mode and the mechanism by which \ac{TOOD} corrects it directly visible.

\paragraph{Uncalibrated Drift.} Without calibration (left panel), the \ac{ID} energy distribution (blue) shifts progressively leftward as tasks accumulate. This leftward shift reflects the confidence gap: gradient updates for new classes inflate current task logits while old task logits shrink in absolute scale. Since the sign-reversed energy score $E(x) = \log \sum_{c} \exp(h_c(x))$ aggregates all logits, the shrinkage of old task logit magnitudes directly reduces $E(x)$ for old task \ac{ID} samples. By Task~4, the \ac{ID} density almost entirely overlaps with both Near \ac{OOD} (coral) and Far \ac{OOD} (orange) distributions. This overlap means no global threshold $\tau$ can separate \ac{ID} from \ac{OOD} samples, which is precisely the AUROC collapse reported in~\Cref{tab:main_results}. Notably, the \ac{OOD} distributions themselves remain approximately stationary throughout the stream, suggesting that the degradation is one sided: it is the \ac{ID} scores that drift toward the \ac{OOD} region, not the reverse.

\paragraph{Restored Separability.} After Robust Anchor normalization (right panel), the \ac{ID} distribution is shifted rightward and stabilized across all task steps, while \ac{OOD} distributions remain anchored at low values. The mechanism behind this \emph{selective} realignment is the per-task decomposition in \Cref{eq:task_energy,eq:robust}. An \ac{ID} sample from task $T_i$ activates the energy channel $E_i$ dominantly; the Robust Anchor normalization (\Cref{eq:robust}) corrects $E_i$ upward using task-specific statistics $(\tilde{e}_i, \widehat{\text{MAD}}_i)$, and the $\max$ operation in \Cref{eq:tood} selects this corrected channel. Under the intended separation, an \ac{OOD} sample typically activates no single channel strongly, so its per-task energies tend to remain low and normalization provides no substantial boost to any channel. The $\max$ over these values therefore tends to yield a low final score. Because the normalization occurs \emph{inside} the $\max$, the effective transformation differs between \ac{ID} and \ac{OOD} samples, enabling \ac{TOOD} to alter AUROC in a way that scalar post-hoc calibration cannot (\Cref{sec:decomp}).

\section{A Controlled Toy Experiment with Known Geometry}
\label{app:abl_toy}

\begin{figure}[!ht]
  \centering
  \includegraphics[width=\linewidth]{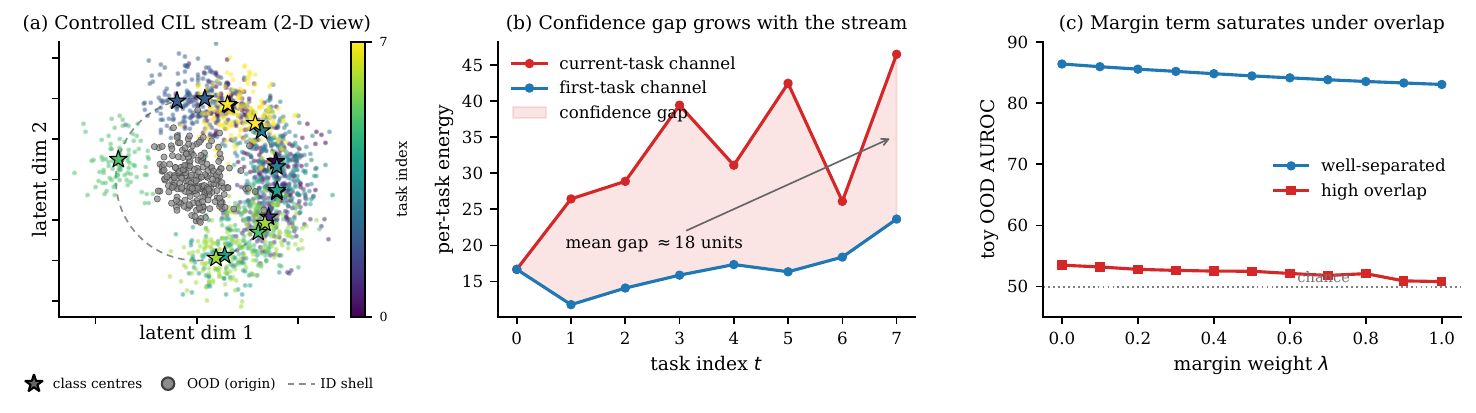}
  \caption{Controlled single-seed confidence-gap toy experiment.
    (a)~the toy geometry: class clusters on a shell
    around a central \ac{OOD} core; (b)~the per-task confidence gap grows
    along the stream ($\approx18$ energy units) and \ac{TOOD} re-centres it;
    (c)~the margin term ($\lambda$) saturates once classes overlap.}
  \label{fig:toy_confidence_gap}
\end{figure}

\textbf{Why a toy experiment.} In a trained network the two mechanisms are entangled:
as the stream grows the classifier head expands (driving the confidence gap)
\emph{and} new classes colonize latent space (driving manifold crowding), so
neither can be switched off to see which one a score-level method actually
repairs. A synthetic stream with a \emph{fixed, fully known} data geometry lets
us drive the two independently---grow the head while holding the data fixed, or
perturb the geometry while holding the head fixed---and read off exactly what
\ac{TOOD} can and cannot recover. The base geometry is $16$ Gaussian-blob
classes ($2$ introduced per task over $8$ tasks) whose centres lie on a sphere
in a $16$-dimensional latent space, with the \ac{OOD} distribution placed as a
blob at the interior origin (\Cref{fig:toy_confidence_gap}, panel a). The
classes are well separated, so in the frozen-geometry regime any detection
collapse is attributable to score drift alone; we introduce crowding only
later, by forcing the blobs to overlap.

\textbf{Protocol.} This is a single-seed diagnostic. A small \ac{MLP} whose
classifier head grows by two units per task is trained sequentially on this
stream and scored with the exact \ac{TOOD} implementation from the main
experiments. The expanding head is the structural cause of the confidence gap
in \ac{CIL}---every task appends logits and rescales the old ones---so
replicating only this growth on top of frozen data reproduces the gap with no
representation change whatsoever. Manipulating the head and the geometry
independently is what isolates the two mechanisms below.

\textbf{Isolating the confidence gap.} As tasks accumulate the current
task's per-task energy drifts well above that of the first task's own
channel---a mean gap of $\approx18$ energy units
(\Cref{fig:toy_confidence_gap}, panel b)---which is precisely the confidence
gap. Because the data geometry is fixed, this drift is purely a score-level
effect, and \ac{TOOD} re-centres the channels, recovering up to $+5.1$ Avg
AUROC over uncalibrated energy across the first five tasks. The toy experiment
therefore exhibits the confidence gap, and \ac{TOOD}'s repair of it, in the
absence of any representation change.

\textbf{Isolating manifold crowding.} To isolate the complementary
mechanism we instead hold the head fixed and \emph{force the class blobs to
physically overlap}, collapsing the geometric margin while leaving the
output-head dynamics benign. Detection now falls to chance
($\approx\!50\%$ AUROC), and \ac{TOOD} provides essentially zero recovery,
because the lost separability lives in the representation geometry rather than
the output scores. The two manipulations thus dissociate the mechanisms
cleanly: a score-level fix recovers the frozen-geometry (confidence-gap)
regime but not the overlap (manifold-crowding) regime, mirroring the
feature-space-analog negative result in \Cref{app:abl_crowding}.

\textbf{Margin failure mode.} The toy experiment also reproduces the margin term's
boundary condition: the optional margin weight $\lambda$ helps while classes
are well separated but saturates and then degrades once they overlap
(\Cref{fig:toy_confidence_gap}, panel c), mirroring the trend in
\Cref{app:ablation_margin}.

\section{Calibration-Component, Margin, Partition, Buffer, and Anchor Ablations}
\subsection{Effect of the margin parameter}
\label{app:ablation_margin}
\begin{figure*}[!ht]
    \centering
    \includegraphics[width=\linewidth ]{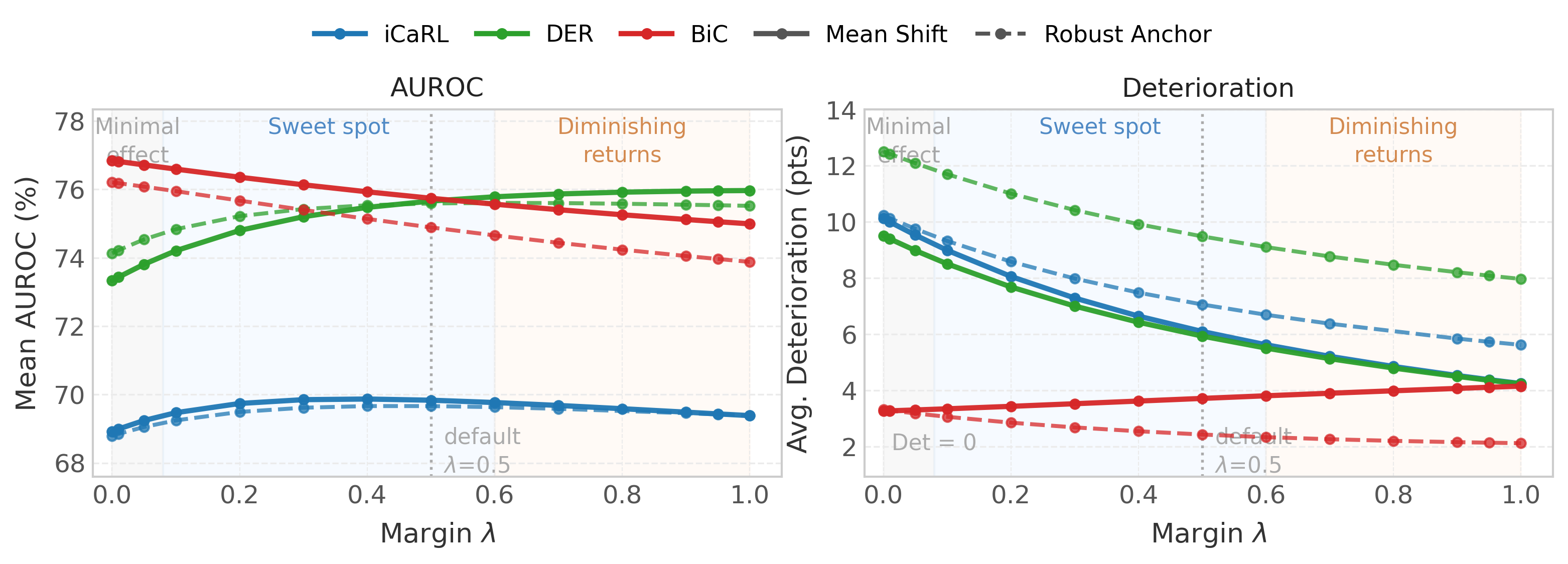}
    \caption{%
        \textbf{Sensitivity to the margin parameter $\lambda$ (CIFAR-10, $N=5$).}
        \emph{Left:} Mean \ac{AUROC} ($\uparrow$); \emph{Right:} Average \ac{OOD} deterioration $D_{\text{avg}}$ ($\downarrow$).
        High-drift methods (iCaRL, DER) peak near $\lambda = 0.5$ (dotted line).
        Low-drift methods (BiC) decrease monotonically for $\lambda > 0$, suggesting that the margin is unnecessary when output head bias correction already stabilizes scores.
    }
    \label{fig:ablation_lambda}
\end{figure*}
The margin term $\lambda$ in~\Cref{eq:margin} augments the base \ac{TOOD} score (\Cref{eq:tood}) by rewarding samples whose top normalized energy $E_{(1)}^{\text{norm}}$ is well separated from the second largest $E_{(2)}^{\text{norm}}$. The intuition is straightforward: an \ac{ID} sample from task $T_i$ produces a dominant energy peak in channel $i$ and weak activations elsewhere, yielding a large gap $E_{(1)}^{\text{norm}} - E_{(2)}^{\text{norm}}$. An \ac{OOD} sample, which belongs to no learned task, produces diffuse, roughly uniform activations across channels, yielding a small gap. The parameter $\lambda$ controls the weight of this discriminative signal. When $\lambda = 0$, \Cref{eq:margin} reduces to the base \ac{TOOD} score; as $\lambda$ increases, large-gap samples receive a larger nonnegative bonus, whereas samples with ambiguous cross-task activation receive a smaller bonus. \Cref{fig:ablation_lambda} evaluates this sensitivity across $\lambda \in [0, 1]$ in $0.1$ increments for three \ac{CL} methods, revealing two distinct behavioral regimes.

\textbf{Regime 1: High-Drift Methods (iCaRL, DER).}
These methods exhibit a large confidence gap (\Cref{sec:gap}), meaning that old task logits have shrunk substantially relative to current task logits by the end of the stream. After \ac{TOOD}'s per-task normalization corrects the location drift, residual noise in the normalized energies can still cause \ac{OOD} samples to achieve moderately high $E_{(1)}^{\text{norm}}$ values through spurious cross-task activations. The margin heuristic addresses these cases indirectly by preferentially boosting samples with a clearly dominant task channel rather than by lowering ambiguous scores. Increasing $\lambda$ from $0$ to $0.5$ yields $+0.9$ AUROC points for iCaRL and $+2.4$ points for DER. Beyond $\lambda = 0.5$, performance saturates for iCaRL and continues to improve slightly for DER, indicating that excessive margin weight can distort the ranking of legitimate \ac{ID} samples whose secondary task activations are non-negligible (e.g., semantically related classes spanning adjacent tasks).

\textbf{Regime 2: Low-Drift Methods (BiC).}
BiC applies explicit bias correction to the output head~\citep{wu2019large}, which already reduces the confidence gap before \ac{TOOD} intervenes. Consequently, the normalized per-task energies for BiC are already well separated, and the base \ac{TOOD} score ($\lambda = 0$) captures the \ac{ID}/\ac{OOD} distinction effectively. Introducing the gap-dependent bonus with $\lambda > 0$ harms the ranking in this regime: the monotonic decline in AUROC (left panel) and increase in $D_{\text{avg}}$ (right panel) show that reweighting by the top-two gap is unnecessary when the underlying score distributions are already calibrated. The optimal value for BiC is $\lambda = 0$.

\textbf{Consensus Value and Practical Guidance.} We set $\lambda = 0.5$ as a fixed default for all main experiments. This value maximizes gains for methods with significant logit drift (the primary use case for \ac{TOOD}) while imposing only modest overhead on methods that do not require calibration. For BiC specifically, the AUROC reduction at $\lambda = 0.5$ is small ($1.1$ point), whereas the gains for DER exceed $+2$ points. In practice, if the \ac{CL} method is known to include output head correction, setting $\lambda = 0$ is preferable; otherwise, $\lambda = 0.5$ provides a robust default without per-method tuning.

\label{app:rebuttal_ablation}

The following analyses isolate the calibration component and the class-to-task
partition used by \ac{TOOD}. All analyses are \emph{post-hoc}: \ac{TOOD} adds
no training and re-scores the same checkpoints used in the main results.
Unless stated otherwise, we
report the change in Avg AUROC (Eq.~\ref{eq:avg_auroc}, the near/far mean)
relative to uncalibrated energy, averaged over the five \ac{CL} methods of
Table~\ref{tab:main_results} and both CIFAR benchmarks;
\Cref{fig:mechanism_dissociation} summarizes the score-level comparisons.

\subsection{The gains come from per-task energy decomposition}
\label{app:abl_mechanism}
\begin{figure}[!ht]
  \centering
  \includegraphics{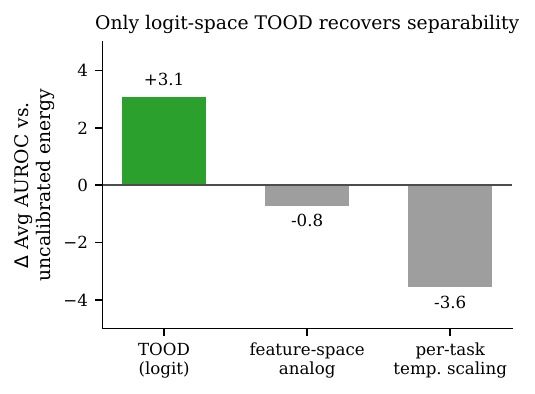}
  \caption{Score-level dissociation.
  Only logit-space \ac{TOOD} ($+3.1$) recovers separability, while generic per-task temperature scaling ($-3.6$) and a feature-space port of the same normalization ($-0.76$) do not.}
  \label{fig:mechanism_dissociation}
\end{figure}

In principle \ac{TOOD} might improve detection simply by being
task-aware---rescaling each task's scores to a common range---rather than
through its particular energy decomposition and reference anchoring. To
separate the two, we add a per-task \emph{temperature-scaling} baseline that
sees the same partition but only rescales each task's logits, without
decomposing or anchoring the energies (\Cref{eq:task_energy,eq:robust}).
Concretely, it divides each task energy of \Cref{eq:task_energy} by a per-task
temperature $T_t$ and scores by the maximum over tasks,
\begin{equation}
    E_t^{\text{ts}}(x) \;=\; \frac{E_t(x)}{T_t},
    \qquad
    S_{\text{ts}}(x) \;=\;
        \max_{t \in \{1,\ldots,N\}} E_t^{\text{ts}}(x),
    \label{eq:temp_scale}
\end{equation}
where each per-task temperature $T_t$ is the standard deviation of that task's
calibration energies,
\[
    T_t \;=\; \sqrt{\frac{1}{|\mathcal{B}_t|}
        \sum_{x' \in \mathcal{B}_t}
        \bigl(E_t(x') - \mu_t\bigr)^2}.
\]
Unlike \Cref{eq:mean_shift,eq:robust}, this baseline has no common reference
recentering (no $\mu_\text{ref}$ or $\tilde{e}_\text{ref}$): each task's
energy is only divided by its own buffer standard deviation and retains its
own offset. It therefore isolates task-aware rescaling without reference
anchoring.
This baseline \emph{lowers} Avg AUROC by $3.6$ points, whereas \ac{TOOD}
raises it by $3.1$ (\Cref{fig:mechanism_dissociation}). Task-aware rescaling
alone is thus insufficient, so the improvement is
attributable to the per-task energy mechanism itself rather than to access to
the partition.

\subsection{What TOOD assumes: the class partition, not task identity}
\label{app:abl_partition}
Because \ac{TOOD} indexes scores by task, one might suspect it covertly
requires task-incremental (\ac{TIL}) inference. It does not, and the two
inputs are different in kind. The \emph{partition} is the fixed training-time
map recording which task each \emph{class} was introduced in; every \ac{CIL}
method already stores it to grow its classifier head, and it is shared by all
test inputs. A \ac{TIL} \emph{task label}, by contrast, is a per-input signal
revealing at inference which task the current sample came from. \ac{TOOD}
uses only the former: it groups the logits into per-task energy channels via
this map and scores an input by the maximum over channels (\Cref{eq:tood}),
so no test sample is ever told its task.

The method is also robust to a \emph{misspecified} partition. To show
this, we set the within-task split factor to three at deployment. Classes are
assigned once to disjoint, balanced subgroups, and empty subgroups are
discarded. This gives three pseudo-groups per 10-class CIFAR-100 task and two
singleton pseudo-groups per 2-class CIFAR-10 task. Averaged across both
datasets, five continual-learning methods, and three seeds, Avg AUROC shifts
by only $-0.5$ points. Detection
therefore tolerates a substantially finer-than-true grouping rather than
requiring accurate task identity at test time, consistent with \ac{TOOD}
operating in the \ac{CIL} regime.

\subsection{Buffer size}

The buffer-size ablation in \Cref{fig:buffer_ablation} shows that performance
stabilizes between $B=50$ and $B=100$, with negligible gains thereafter.
Random class-balanced samples also match iCaRL-style herding to within $0.02$
Avg AUROC points on average.

\subsection{Anchor sensitivity}
\label{app:abl_anchor}

The robust-anchor normalization (\Cref{eq:robust}) maps every task's energy
onto a common reference $(\tilde{e}_\text{ref},
\widehat{\text{MAD}}_\text{ref})$. We default to the most-recent task $T_N$,
but the choice applies the same affine map to all normalized task channels and
therefore does not change their ranking. Across five \ac{CL} methods, both
CIFAR datasets, and three seeds, the largest absolute Avg AUROC changes
relative to the default are $0.0001$ for the oldest-task anchor, $0.0000$ for
the max-spread anchor, and $0.00015$ for the mean-statistics anchor. We retain
$T_N$ only as an interpretable current-scale reference for absolute decision
thresholds.

\section{ImageNet and ViT Scalability}

\subsection{Scalability on ImageNet-1K}
\label{app:imagenet_scalability}

On ImageNet-1K ($N = 100$), \ac{TOOD} (Mean Shift) substantially helps
drift-sensitive methods such as BiC and DER, while WA changes only marginally,
consistent with its stable output scale.

\begin{table}[H]
\centering\small
\setlength{\tabcolsep}{4pt}
\caption{%
    \textbf{ImageNet-1K results ($N=100$; single seed).} Columns are CL methods; rows are OOD detectors. We report average CIL accuracy, Avg AUROC~(\Cref{eq:avg_auroc}) ($\uparrow$), and Avg FPR@95 ($\downarrow$), with detection metrics averaged over Near-OOD and Far-OOD. Bold, underline, and teal denote 1st, 2nd, and 3rd place within each metric column.}
\label{tab:imagenet_results}
\begin{tabular}{l cc cc cc}
\toprule
 & \multicolumn{6}{c}{\textbf{ImageNet-1K ($N = 100$)}} \\
\cmidrule(lr){2-7}
\textbf{OOD Method} & \multicolumn{2}{c}{BiC} & \multicolumn{2}{c}{WA} & \multicolumn{2}{c}{DER} \\
 & AUC $\uparrow$ & FPR $\downarrow$ & AUC $\uparrow$ & FPR $\downarrow$ & AUC $\uparrow$ & FPR $\downarrow$ \\
\midrule
Avg CIL Acc. ($\uparrow$) & \multicolumn{2}{c}{28.9} & \multicolumn{2}{c}{26.8} & \multicolumn{2}{c}{30.8} \\
\midrule
MSP &  57.7  &  86.3  & 62.8 & 82.5 & 62.4 & 84.6 \\
ODIN & 56.4 & 89.5 & 63.9 & 82.0 & 66.0 & 81.2 \\
Ash & 56.3 & 91.3 & \underline{67.5} & 81.6 & 68.7 & 80.4 \\
ADASCALE (Activation) & 44.8 & 95.0 & 58.1 & 88.1 & 54.1 & 93.4 \\
NNGuide & 42.0 & 92.7 & 55.4 & 86.5 & 47.0 & 91.9 \\
Energy (Uncalib.) & \textcolor{teal}{59.2} & \textcolor{teal}{86.1} & \textbf{70.1} & \textbf{76.4} & \textcolor{teal}{70.2} & \underline{76.9} \\
\midrule
Energy + Mean Shift & \textbf{65.2} & \underline{76.5} & \textcolor{teal}{65.7} & \underline{77.4} & \textbf{71.4} & \textbf{76.2} \\
Energy + Robust Anchor & \underline{65.1} & \textbf{75.2} & 65.6 & \textcolor{teal}{79.0} & \underline{70.5} & \textcolor{teal}{77.8} \\
\bottomrule
\end{tabular}
\end{table}

\subsection{Vision-Transformer backbone}
\label{app:abl_vit}
Our main results use ResNet backbones, whose BatchNorm statistics could in
principle drive the confidence gap. To test a different normalization regime,
we ran single-seed experiments with a ViT-B/16 backbone (ImageNet-pretrained,
LayerNorm) on the
CIFAR-10 five-task stream and re-scored the checkpoints post hoc. The correction
transfers across the normalization regime: for two methods, \ac{TOOD}
raises Avg AUROC over uncalibrated energy by $+3.5$ points on iCaRL
($54.5 \to 58.0$) and $+1.1$ on WA, matching the direction and magnitude of the
ResNet results and indicating the fix is not BatchNorm-specific.

On CIFAR-100, where the confidence gap is most severe, the gains are
larger (\Cref{tab:vit_cifar100}). \ac{TOOD} improves WA by $7.4$ Avg AUROC
points ($60.9$ to $68.3$) and BiC by $1.0$, while DER ($-0.1$) and LwF ($-0.8$)
are left essentially unchanged. That WA improves more on CIFAR-100 than on
CIFAR-10 fits the picture of a gap that widens as the stream grows longer and
the number of classes increases. As on ResNet (\Cref{sec:setup}) the effect
varies by method, because \ac{TOOD} is a targeted repair for score drift: it
helps where a confidence gap is present and stays neutral, or turns slightly
negative, where it is not.

These results also help assess whether the gain merely follows the imbalance of
accuracy across tasks. WA and DER both collapse onto the most recent task,
reaching roughly $86\%$ accuracy there against at most $30\%$ on the older
tasks, yet \ac{TOOD} moves them in opposite directions, by $+7.4$ for WA and
$-0.1$ for DER. Two models with almost identical imbalance thus behave very
differently, suggesting that the benefit follows the confidence gap in
the output scores rather than the balance of accuracy across tasks. We report
these methods as a single-seed proof of concept and leave a fuller multi-seed
ViT study to future work.

\begin{table}[!ht]
\centering\small
\caption{\textbf{ViT-B/16 CIFAR-100 (10-task), post-hoc, single-seed.}
Avg CIL accuracy (as in Table~\ref{tab:main_results}) and Avg AUROC (near+far)
under uncalibrated energy vs.\ \ac{TOOD} (robust anchor).}
\label{tab:vit_cifar100}
\begin{tabular}{lcccc}
\toprule
Method & Avg CIL Acc. & Uncalib. & \ac{TOOD} & $\Delta$ \\
\midrule
WA  & $35.6$ & $60.9$ & $68.3$ & $\mathbf{+7.4}$ \\
BiC & $54.5$ & $65.6$ & $66.6$ & $+1.0$ \\
DER & $47.5$ & $62.8$ & $62.7$ & $-0.1$ \\
LwF & $46.4$ & $64.7$ & $63.9$ & $-0.8$ \\
\bottomrule
\end{tabular}
\end{table}

\end{document}